\documentclass[table]{article}

\usepackage[nonatbib, preprint]{neurips_data_2022}

\usepackage[square,numbers]{natbib}
\usepackage[utf8]{inputenc} 
\usepackage[T1]{fontenc}    
\usepackage{url}            
\usepackage{booktabs}       
\usepackage{amsfonts}       
\usepackage{nicefrac}       
\usepackage{microtype}      
\usepackage{array,multirow}
\usepackage{subcaption}
\usepackage{caption}
\usepackage{xspace}
\usepackage{multirow}
\usepackage{enumitem}
\usepackage{graphicx}
\usepackage{wrapfig}
\usepackage{float}
\usepackage{booktabs}
\usepackage{hyperref}
\hypersetup{
    colorlinks=true,
    urlcolor=cyan
    }
\usepackage{tikz}
\usepackage{collcell}
\usepackage{amssymb}
\usepackage{pifont}
\usepackage{amsmath}
\usepackage{soul}

\newcommand{\colorfromval}[3]{%
    \pgfmathsetmacro{\PercentColor}{100.0*(#3-#1)/(#2-#1)}
    \pgfmathparse{#3>=#2?int(0):int(1)}
    \ifnum\pgfmathresult>0\relax 
        {\textcolor{black!\PercentColor!gray}{#3}}
    \else
        \pgfmathparse{#3>=0?int(1):int(0)}
        \ifnum\pgfmathresult>0\relax 
            \textbf{\textcolor{black!\PercentColor!gray}{#3}}
        \else 
             {\textcolor{gray}{NA}}
        \fi
    \fi
}
\newcolumntype{T}[2]{>{\collectcell{\colorfromval{#1}{#2}}} l <{\endcollectcell}}

\newcommand{\method}{EdgeBank\xspace}

\newcommand{\tableRef}[1]{Table~\ref{#1}}
\newcommand{\figureRef}[1]{Fig.~\ref{#1}}

\newcommand{\Wikipedia}{Wikipedia\xspace}
\newcommand{\Reddit}{Reddit\xspace}
\newcommand{\MOOC}{MOOC\xspace}
\newcommand{\LastFM}{LastFM\xspace}
\newcommand{\Enron}{Enron\xspace}
\newcommand{\SocialEvo}{Social Evo.\xspace}
\newcommand{\UCI}{UCI\xspace}
\newcommand{\Flights}{Flights\xspace}
\newcommand{\CanParl}{Can. Parl.\xspace}
\newcommand{\USLegis}{US Legis.\xspace }
\newcommand{\UNTrade}{UN Trade\xspace }
\newcommand{\UNVote}{UN Vote\xspace}
\newcommand{\contact}{Contact\xspace}

\newcommand{\TEA}{TEA\xspace}
\newcommand{\TET}{TET\xspace}

\newcommand{\changed}[1]{\textcolor{black}{#1}}

\definecolor{q1}{rgb}{1, 1, 1}  
\definecolor{q2}{HTML}{feebe2}
\definecolor{q3}{HTML}{fbb4b9}
\definecolor{q4}{HTML}{f768a1}

\author{
Farimah Poursafaei\thanks{Equal contribution.} , Shenyang Huang\footnotemark[1] , Kellin Pelrine, Reihaneh Rabbany \\
McGill University School of Computer Science, Mila – Quebec AI Institute  \\
\lbrack farimah.poursafaei,huangshe,kellin.pelrine,reihaneh.rabbany\rbrack @mila.quebec \\
}

\title{Towards Better Evaluation for \\ Dynamic Link Prediction}

\begin{document}

\maketitle

\begin{abstract}

Despite the prevalence of recent success in learning from static graphs, learning from time-evolving graphs remains an open challenge. In this work, we design new, more stringent evaluation procedures for link prediction specific to dynamic graphs, which reflect real-world considerations, to better compare the strengths and weaknesses of methods. First, we create two visualization techniques to understand the reoccurring patterns of edges over time and show that many edges reoccur at later time steps. Based on this observation, we propose a pure memorization baseline called \method. \method achieves surprisingly strong performance across multiple settings because easy negative edges are often used in current evaluation setting. To evaluate against more difficult negative edges, we introduce two more challenging negative sampling strategies that improve robustness and better match real-world applications. Lastly, we introduce six new dynamic graph datasets from a diverse set of domains missing from current benchmarks, providing new challenges and opportunities for future research. 
\changed{Our code repository is accessible at \url{https://github.com/fpour/DGB.git}.}

\end{abstract}

\section{Introduction}
\label{sec:introduction}

Many evolving real-world relations can be modelled by a dynamic graph where nodes correspond to entities and edges represent relations between nodes. Nodes, edges, weights or attributes in a dynamic graph can be added, deleted or adjusted over time. Therefore, understanding and analyzing the temporal patterns of a dynamic graph is an important problem. 
For instance, in popular online social networks, many users join the platform on a daily basis while connections between users are constantly added or removed~\cite{haghani2019systemic}. 
To facilitate more efficient learning on dynamic graphs, many efforts have been devoted to the development of dynamic graph representation learning methods~\cite{wang2021adaptive, tian2021streaming,wang2020inductive,rossi2020temporal,xu2020inductive,yang2021discrete,sankar2020dysat,bielak2022fildne}.

\begin{wrapfigure}{R}{0.5\textwidth}\vspace{-14pt}
\centering
\includegraphics[width=0.5\textwidth]{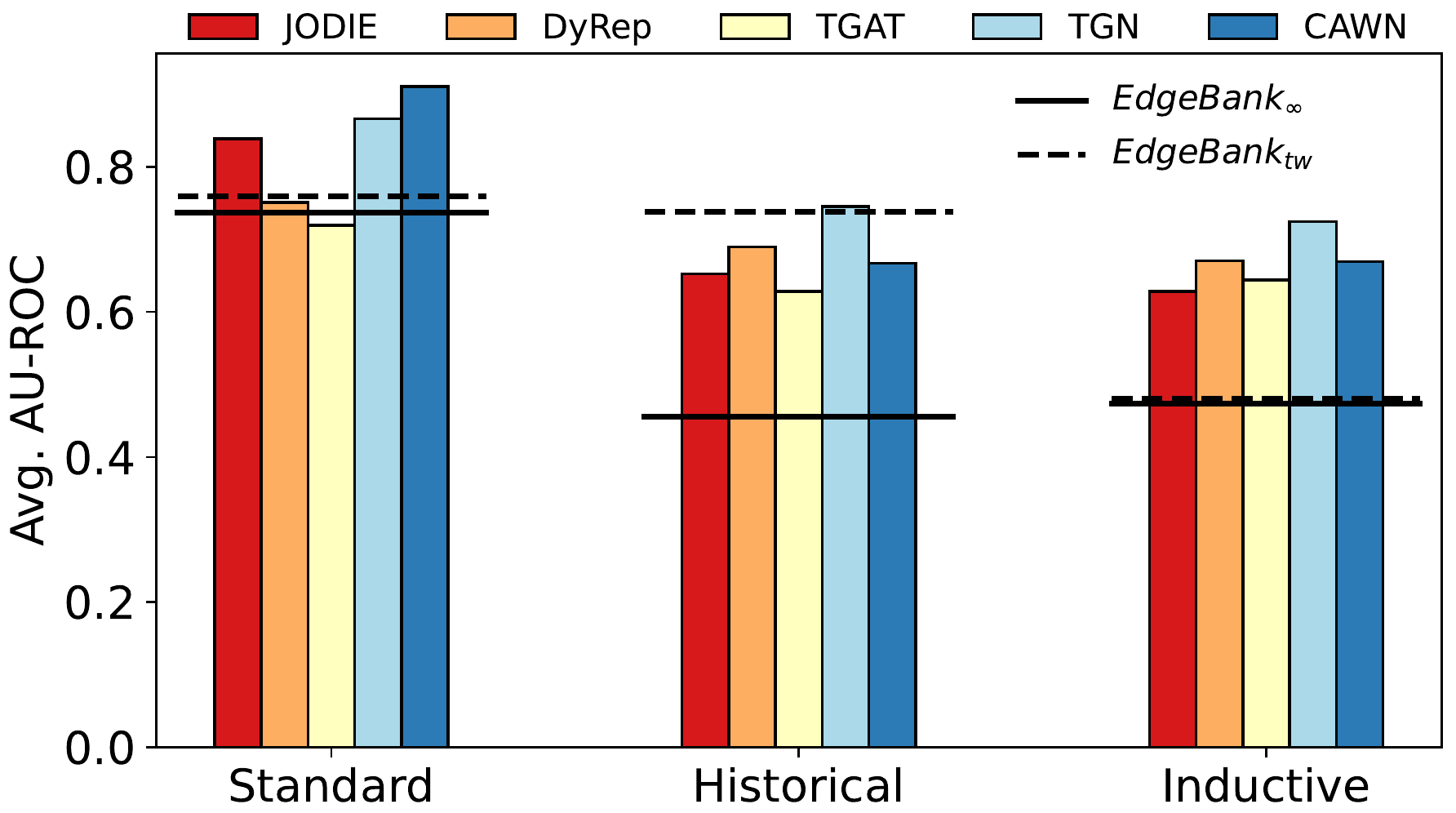}
\caption{The ranking of different methods changes in the proposed negative sampling settings \changed{which contains more difficult negative edges.} Our proposed baselines (horizontal lines) show competitive performance, in particular in the standard setup.
\changed{The results illustrate the average performance over all datasets presented in~\tableRef{tab:stats}.}
}
\label{fig:diff_ns}\vspace{-12pt}
\end{wrapfigure}

Link prediction is a fundamental learning task on dynamic graphs which focuses on predicting future connections between nodes.
Recent methods such as~\cite{kumar2019predicting, trivedi2019dyrep, xu2020inductive, rossi2020temporal, wang2020inductive} show promising performance on this task, with
the state-of-the-art~(SOTA) performance~\cite{rossi2020temporal, wang2020inductive} being close to perfect on most existing benchmark datasets. 
However, considering that link prediction in static graphs, an arguably less complex task, still faces major challenges~\cite{hu2020open, hu2021ogb}, it is important to meticulously examine the near-perfect performance of dynamic link prediction methods.
We hypothesize that current evaluation procedures and datasets fail to properly differentiate between the proposed approaches. Therefore, we identify several limitations in the current evaluation procedure and propose solutions towards more robust and effective evaluation protocols.

\textbf{Limited Domain Diversity.}
Existing benchmark datasets are mostly social or interaction networks thus limited in domain diversity. It is well-known that networks across different domains exhibit a diverse set of properties. For example, biological networks such as protein interaction networks differ significantly from social networks in community structure and centrality measures~\cite{gutierrez2014biological}. Therefore, it is necessary to test dynamic link prediction methods in various domains outside of social or interaction networks. To this end, we incorporate six new datasets for dynamic link prediction ranging from politics, economics, and transportation networks. In addition, we introduce novel visualization techniques for dynamic graphs. We show that in most networks, a significant portion of edges reoccur over time but the reocurrence patterns vary widely across different networks and domains.

\textbf{Easy Negative Edges.} \changed{In a dynamic network, the edges that have been never observed during previous timestamps can be considered as \textit{easy} negative edges, since it is less likely that these edges occur during the test phase given the reoccurring pattern of dynamic graphs.} We introduce two novel Negative Sampling~(NS) strategies, specifically designed to incorporate more difficult edges in dynamic graphs, which select negative edges based on the reocurrence of observed edges. As shown in \figureRef{fig:diff_ns}, SOTA methods have a significant decrease in performance when a different set of negative edges is sampled during test time. Moreover, the relative ranking of methods varies significantly across NS settings. Thus, it is important to evaluate methods on different sets of negative edges.

\textbf{Memorization Works Well.}
Finally, we introduce a simple memorization-based baseline, named \method, which simply stores previously observed edges in memory, and then predicts existing edges in memory as positive at test time. In \figureRef{fig:diff_ns}, we contrast the performance of SOTA methods with that of \method~(in horizontal lines). \method is a surprisingly strong baseline for dynamic link prediction. In the historical NS setting, \method achieves the second best ranking amongst all methods. As \method requires neither learning nor hyper-parameter tuning, we argue that it is a strong and necessary baseline for future methods to compare against.

The goal of this work is to propose more effective evaluation strategies to better differentiate dynamic link prediction methods. We identify challenges and drawbacks in the current evaluation setting for dynamic link prediction: (1) existing strategies for sampling negative edges during evaluation are insufficient, (2) memorization leads to over-optimistic evaluation, and (3) there is a lack of diversity in dynamic graph dataset domains. Our main contributions can be summarized as follows:
\begin{itemize}[leftmargin=5pt,topsep=0pt, noitemsep]
    
    \item \textbf{Novel Negative Sampling Strategies}. We evaluate the impact of negative edges on model performance and outline two novel sampling strategies: \emph{historical NS} and \emph{inductive NS}, which provide more robust and in depth evaluation.
    
    \item \textbf{Strong Baseline}. We propose a novel non-parameterized and memorization-based method, \method, which provides a strong baseline for current and future approaches to compare against. 
  
    \item \textbf{New Datasets and Visualization Tools}. We present six novel dynamic graph datasets from various domains such as politics, transportation, and economics. These datasets exhibit different temporal edge evolution patterns, which can be understood through our proposed \TEA and \TET plots.   
    
    
    
\end{itemize}

\textbf{Reproducibility}: our code repository is available at \url{https://github.com/fpour/DGB.git}. 
\changed{All datasets can be accessed at \url{https://zenodo.org/record/7008205\#.Yv_a_3bMJPZ}.}








\section{Related Work} \vspace{-7pt}

\textbf{Benchmarking Graph Learning Methods.} A number of studies identify several issues in evaluation of existing GNN models \cite{dwivedi2020benchmarking,shchur2018pitfalls,errica2020fair,hu2020open,lv2021we}. 
Focusing on 
static graphs, 
Dwivedi et al. \cite{dwivedi2020benchmarking} identify issues with comparative evaluation due to inconsistent experimental settings.
Shchur et al. \cite{shchur2018pitfalls} 
show that reusing the same train-test splits in many different works has led to overfitting and using different splits of the data could result in different ranking of the methods.
OGB~\cite{hu2020open} facilitates reproducibility and scalability of graph learning tasks by providing a diverse set of datasets together with unified evaluation protocols, metrics, and data splits.
In contrast to these works, we focus on improving evaluation for \textit{dynamic} link prediction.


For dynamic graphs, Junuthula et al. \cite{junuthula2016evaluating} differentiate dynamic and static link prediction by edge insertion or deletion. 
Junuthula et al. \cite{junuthula2018leveraging} then consider the problem of incorporating information from friendship networks into predicting future links in social interaction domains. 
Haghani and Keyvanpour \cite{haghani2019systemic} provide a comprehensive review of link prediction methods for social networks and categorize the link prediction task into two groups: missing link prediction, and future link prediction.
Similar to these works, we also focus on dynamic link prediction but from different perspectives: new negative sampling strategies, new baseline, and new dataset domains.

\textbf{Negative Sampling (NS) of Edges in Graphs.}
Yang et al. \cite{yang2020understanding} argue that NS is as important as positive sampling in graph representation learning.
For static link prediction, the most common method is to sample negative edges at random \cite{grover2016node2vec,backstrom2011supervised,scripps2008matrix}. 
Alternatively, the sampling can be based on connecting nodes with specific properties (e.g. a sufficiently large degree) \cite{liben2007link}, or it can be based on a particular geodesic distance \cite{lichtenwalter2010new,lu2011link}. 
Kotnis and Nastase \cite{kotnis2017analysis} provide an empirical study of the impact of different NS strategies during training on the learned representations of various methods in knowledge graphs.
In our work, we focus on the impact of NS strategies during evaluation, and propose two novel NS strategies based on the history of the observed edges in dynamic graphs. Current evaluation protocol has difficulty differentiating between models as many methods achieve near-perfect performance across the board. In comparison, our proposed NS strategies sample harder negative edges for better evaluation.

\textbf{Dynamic Graph Representation Learning.}
Recently there is a surge of interest towards temporal networks. 
Kazemi et al. \cite{kazemi2020representation} present a survey of advances in representation learning on dynamic graphs.
Skardinga et al. \cite{skardinga2021foundations} concentrate on recent studies on Dynamic Graph Neural Networks (DGNNs) and provide a detailed terminology of dynamic networks. Zhang et al.~\cite{zhang2021cope} highlights the importance of learning \textit{fully} temporal embeddings which also models information propagation.  
Skardinga et al. \cite{skardinga2021foundations} and  Kazemi et al. \cite{kazemi2020representation} both argue modeling dynamic graphs with continuous representations has higher potential, since it offers superior temporal granularity. 
In our experiments we center our attention on five recent models of this type: 
\textit{JODIE} \cite{kumar2019predicting}, \textit{DyRep} \cite{trivedi2019dyrep}, \textit{TGAT} \cite{xu2020inductive}, \textit{TGN} \cite{rossi2020temporal}, and \textit{CAWN} \cite{wang2020inductive}.
We summarize these methods in Appendix \ref{subsec:baseline_methods}.
As shown in the experiments section, these methods often achieve close to perfect performance for current link prediction tasks on dynamic graphs. This hinders researchers' ability to evaluate if new models are superior. Also, it exaggerates the efficacy of current models on real-world tasks. 
Hence, we further examine the evaluation procedure, from the perspective of both benchmark datasets and negative sampling.

\section{Understanding Dynamic Graph Datasets}
\label{sec:dataset}

A dynamic graph can be represented as timestamped edge streams -- triplets  of source, destination, timestamp, i.e. $\mathcal{G}=\{ (s_1,d_1,t_1), (s_2,d_2,t_2), \dots \} $ where the timestamps are ordered (\textit{$0 \leq t_1 \leq t_2 \leq ... \leq T $}).
We investigate the task of predicting the existence of an edge between a node pair in the future. The timeline is split at a point, $t_{split}$, into all edges appearing before or after. This results in train and test edge sets $E_\text{train}$ and $E_\text{test}$. We can then divide edges of a given dynamic graphs into three categories:
(a) edges that are only seen during training ($E_\text{train} \setminus{E_\text{test}}$), 
(b) edges that are seen during training and reappear during test ($E_\text{train} \cap {E_\text{test}}$), which can be considered as \textit{transductive} edges, and 
(c) edges that have not been seen during training and only appear during test ($ {E_\text{test}}\setminus{E_\text{train}}$), which can be considered as \textit{inductive} edges.

\begin{table*}[t]\vspace{-12pt}
\caption{Dataset statistics.}\vspace{-5pt}
\label{tab:stats}
  \resizebox{\linewidth}{!}{%
  \begin{tabular}{l | l l l l l l l}
  \toprule
  \toprule
  Dataset & Domain & \# Nodes & Total Edges
  & Unique Edges & Unique Steps & Time Granularity & Duration \\ 
  \midrule

  \Wikipedia & Social & 9,227 & 157,474 & 18,257 & 152,757 & Unix timestamp & 1 month \\ 
  \Reddit & Social & 10,984 & 672,447 & 78,516 & 669,065 & Unix timestamp & 1 month \\ 
  \MOOC & Interaction & 7,144 & 411,749 & 178,443 & 345,600 & Unix timestamp & 17 month \\ 
  \LastFM & Interaction & 1,980 & 1,293,103 & 154,993 & 1,283,614 & Unix timestamp & 1 month \\ 
  \Enron & Social & 184 & 125,235 & 3,125 & 22,632 & Unix timestamp & 3 years \\ 
  \SocialEvo & Proximity & 74 & 2,099,519 & 4,486 & 565,932 & Unix timestamp & 8 months \\ 
  \UCI & Social & 1,899 &59,835 & 20,296 & 58,911 & Unix timestamp & 196 days \\ 
  
  \midrule
  
  \Flights (new) & Transport & 13,169 & 1,927,145 & 395,072 & 122 & days & 4 months \\ 
  \CanParl (new) & Politics & 734 & 74,478 & 51,331 & 14 & years & 14 years \\ 
  \USLegis (new) & Politics & 225 & 60,396 & 26,423 & 12 & congresses & 12 congresses \\ 
  \UNTrade (new) & Economics & 255 & 507,497 & 36,182 & 32 & years & 32 years \\ 
  \UNVote (new) & Politics & 201 & 1,035,742 & 31,516 & 72 & years & 72 years \\ 
  \changed{\contact (new)} & \changed{Proximity} & \changed{694} & \changed{2,426,280} & \changed{79,531} & \changed{8,065} & \changed{5 minutes} & \changed{1 month} \\
  \bottomrule
  \bottomrule
  \end{tabular}
  }
\vspace{-10pt}
\end{table*}

We aim to understand the differences between dynamic graph datasets across a variety of domains. To this end, we first investigate seven widely used benchmark datasets and contribute six novel dynamic graphs~(marked as \textit{new}) from diverse domains currently under-studied in dynamic link prediction literature. The statistics of these datasets are summarized in \tableRef{tab:stats}, and details are explained in Section~\ref{subsec:datasets}. To better characterize differences between dynamic graphs, we propose two types of plots and define three indices to visualize and quantify the patterns in dynamic graphs and the difficulty of a given evaluation split in Section~\ref{subsec:tea} and Section~\ref{subsec:tet}.

\subsection{Temporal Graph Datasets}
\label{subsec:datasets}

\changed{We consider a wide set of dynamic graph datasets from diverse domains. 
None of these datasets contains node attributes, but we include description of edge attributes when applicable. The data collection and processing details are explained in Appendix~\ref{subsec:collection}.} All datasets are publicly available under MIT licence or Apache License 2.0.
\begin{itemize}[align=parleft,left=0pt..1em,topsep=0pt, noitemsep]
    \item \href{http://snap.stanford.edu/jodie/wikipedia.csv}{\textbf{\Wikipedia}~\cite{kumar2019predicting}}: consists of edits on Wikipedia pages over one month. Editors and wiki pages are modelled as nodes, and the timestamped posting requests are edges. \changed{Edge features are LIWC-feature vectors~\cite{pennebaker2001linguistic} of edit texts with a length of 172}.
    
    \item \href{http://snap.stanford.edu/jodie/reddit.csv}{\textbf{\Reddit}~\cite{kumar2019predicting}}: models subreddits' posted spanning one month, where the nodes are users or posts and the edges are the timestamped posting requests. \changed{Edge features are LIWC-feature vectors~\cite{pennebaker2001linguistic} of edit texts with a length of 172}.
    
    \item \href{http://snap.stanford.edu/jodie/mooc.csv}{\textbf{\MOOC}~\cite{kumar2019predicting}}: is a student interaction network formed from online course content units such as problem sets and videos. Each edge is a student accessing a content unit and \changed{has 4 features.}
    
    \item \href{http://snap.stanford.edu/jodie/lastfm.csv}{\textbf{\LastFM}~\cite{kumar2019predicting}}: is an interaction network where users and songs are nodes and each edge represents a user-listens-to-song relation. 
    The dataset consists of the relations of 1000 users listening to the 1000 most listened songs over a period of one month. \changed{The dataset contains no attributes.}
    
    \item \href{https://www.cs.cmu.edu/~./enron/}{\textbf{\Enron}~\cite{shetty2004enron}}: is an email correspondence dataset containing around 50K emails exchanged among employees of the ENRON energy company over a three-year period. \changed{This dataset has no attributes.}
    
    \item \href{http://realitycommons.media.mit.edu/socialevolution.html}{\textbf{\SocialEvo}~\cite{madan2011sensing}}: is a mobile phone proximity network which tracks the everyday life of a whole undergraduate dormitory from October 2008 to May 2009.  \changed{Each edge has 2 features.}
    
    \item \href{http://konect.cc/networks/opsahl-ucforum/}{\textbf{\UCI}~\cite{panzarasa2009patterns}}: is a Facebook-like, \changed{unattributed} online communication network among students of the University of California at Irvine, along with timestamps with the temporal granularity of seconds.
    
    \item \href{https://zenodo.org/record/3974209/#.Yf62HepKguU}{\textbf{\Flights}~(\textit{new})~\cite{schafer2014bringing}}: is a directed dynamic flight network illustrating the development of the air traffic during the COVID-19 pandemic. It was extracted and cleaned for the purpose of this study.  
    Each node represents an airport and each edge is a tracked flight. \changed{The edge weights specify the number of flights between two given airports in a day.}
    
    \item \href{https://github.com/shenyangHuang/LAD}{\textbf{\CanParl}~(\textit{new})~\cite{huang2020laplacian}}: is a dynamic political network documenting the interactions between Canadian Members of Parliaments~(MPs) from 2006 to 2019. 
    Each node is one MP representing an electoral district and each edge is formed when two MPs both voted \textit{"yes"} on a bill. \changed{The edge weights specify the number of times that one MP voted \textit{"yes"} for another MP in a year.}
    
    \item \href{https://github.com/shenyangHuang/LAD}{\textbf{\USLegis}~(\textit{new})~\cite{fowler2006legislative,huang2020laplacian}}: is a senate co-sponsorship graph which documents social interactions between legislators from the US Senate. \changed{The edge weights specify the number of times two congress persons have co-sponsored a bill in a given congress.} 
    
    \item \href{https://www.fao.org/faostat/en/#data/TM}{\textbf{\UNTrade}~(\textit{new})~\cite{macdonald2015rethinking}}: is a weighted, directed, food and agriculture trading graph between 181 nations and spanning over 30 years. 
    \changed{The edge weights specify the total sum of normalized agriculture import or export values between two countries.} 
    
    \item \href{https://dataverse.harvard.edu/dataset.xhtml?persistentId=doi:10.7910/DVN/LEJUQZ}{\textbf{\UNVote}~(\textit{new})~\cite{LEJUQZ_2009}}: is a dataset of roll-call votes in the United Nations General Assembly from 1946 to 2020. 
    \changed{If two nations both voted \textit{"yes"} for an item, then the edge weight between them is incremented by one.}
    
    \item \href{https://springernature.figshare.com/articles/dataset/Metadata_record_for_Interaction_data_from_the_Copenhagen_Networks_Study/11283407/1}{\textbf{\contact}~(\textit{new})~\cite{sapiezynski2019interaction}}: \changed{is a dataset describing the temporal evolution of the physical proximity around 700 university students over a period of four weeks. Each participant is assigned an unique ID and edges between users indicate that they are within close proximity of each other. The edge weights indicate the physical proximity between participants.}
\end{itemize}

\begin{figure*}[t]
\centering
\begin{subfigure}{0.25\textwidth}
 \includegraphics[width=\textwidth]{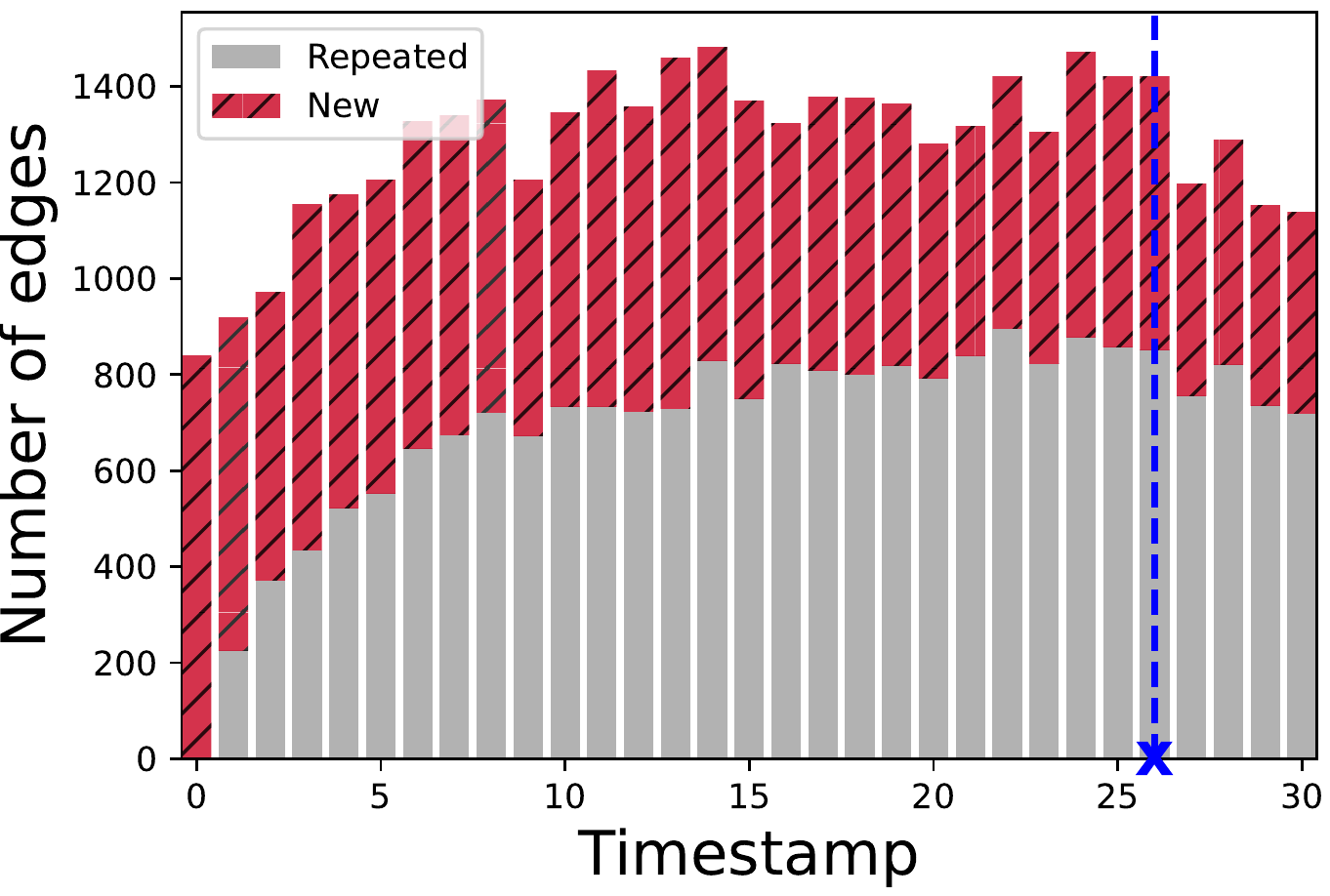}
  \caption{\Wikipedia (0.46)}
\end{subfigure}%
\begin{subfigure}{0.25\textwidth}
 \includegraphics[width=\textwidth]{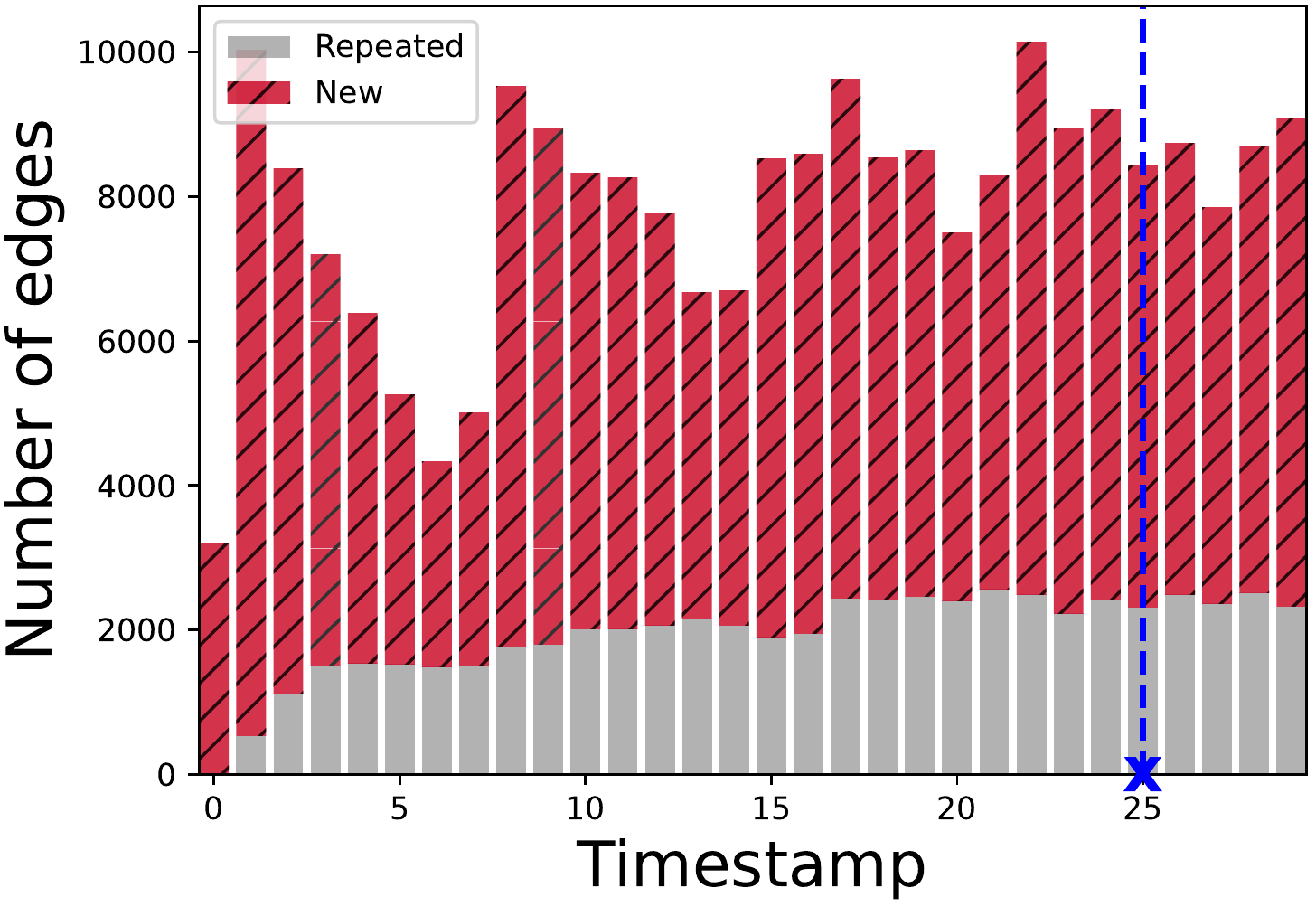}
  \caption{\MOOC (0.75)}
\end{subfigure}%
\begin{subfigure}{0.25\textwidth}
  \centering
  \includegraphics[width=\textwidth]{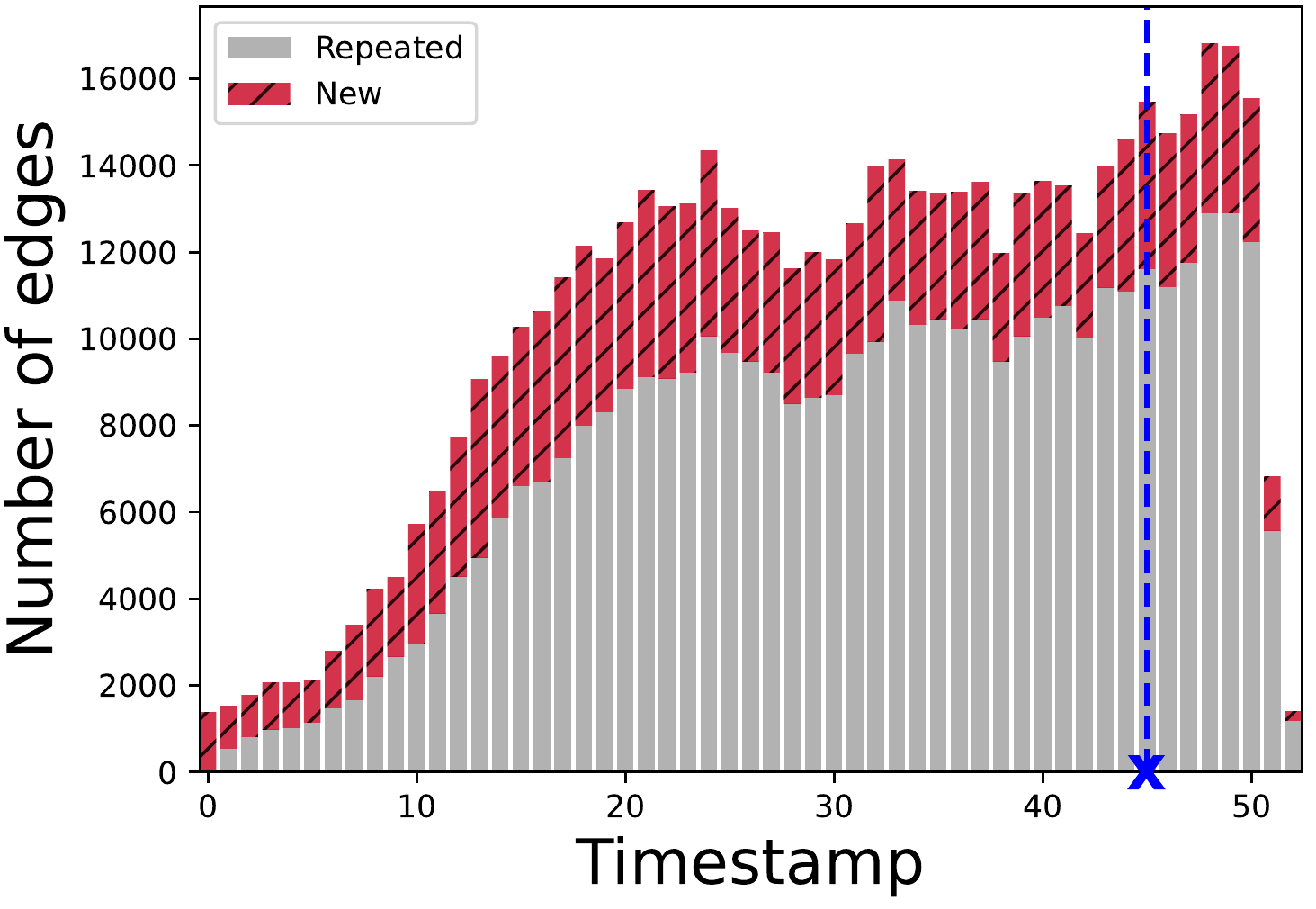}
  \caption{\LastFM (0.28)}
\end{subfigure}%
\begin{subfigure}{0.25\textwidth}
 \includegraphics[width=\textwidth]{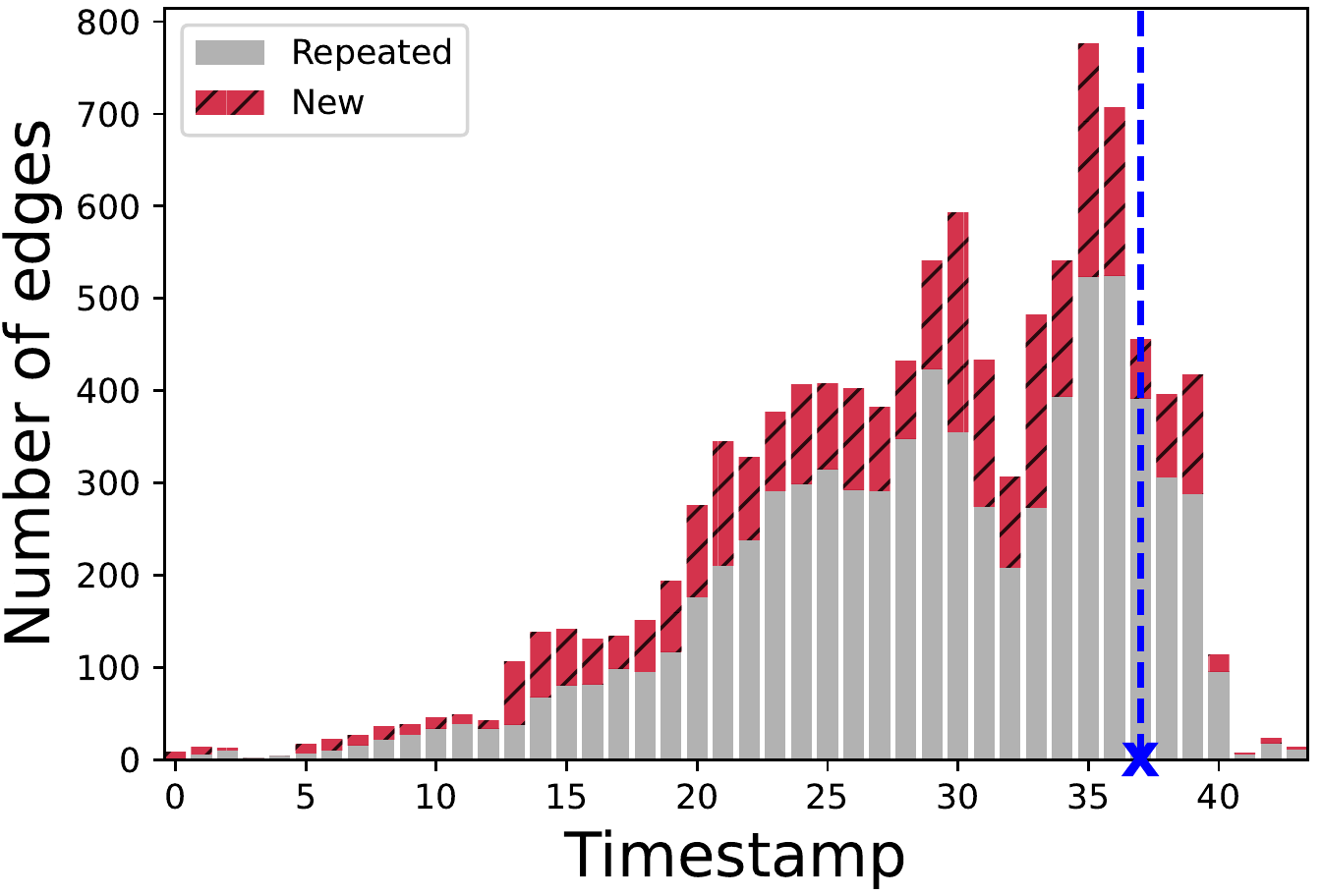}
  \caption{\Enron (0.30)}
\end{subfigure}%

\begin{subfigure}{0.25\textwidth}
  \centering
  \includegraphics[width=\textwidth]{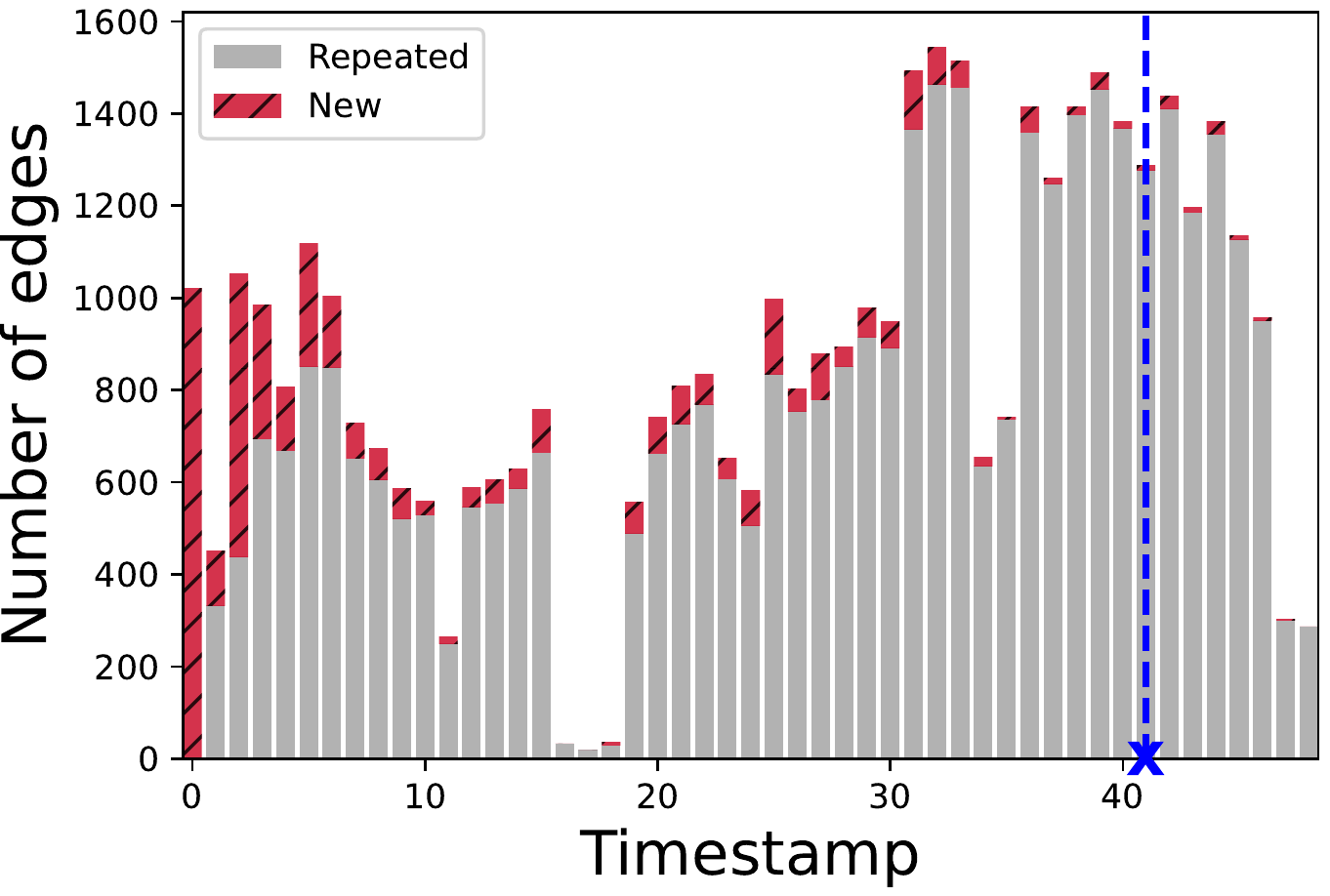}
  \caption{\SocialEvo (0.11)}
\end{subfigure}%
\begin{subfigure}{0.25\textwidth}
 \includegraphics[width=\textwidth]{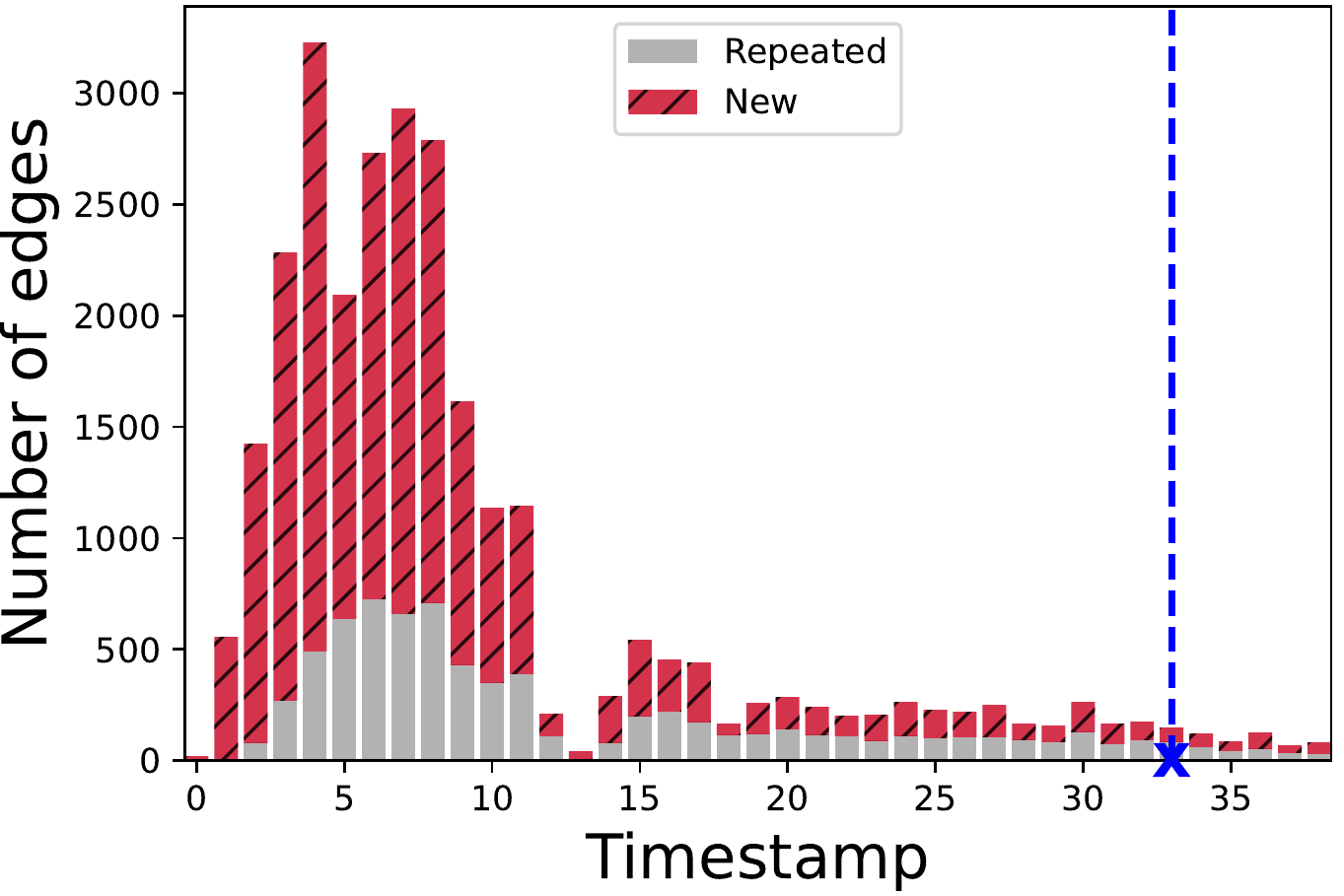}
  \caption{\UCI (0.73)}
\end{subfigure}%
\begin{subfigure}{0.25\textwidth}
 \includegraphics[width=\textwidth]{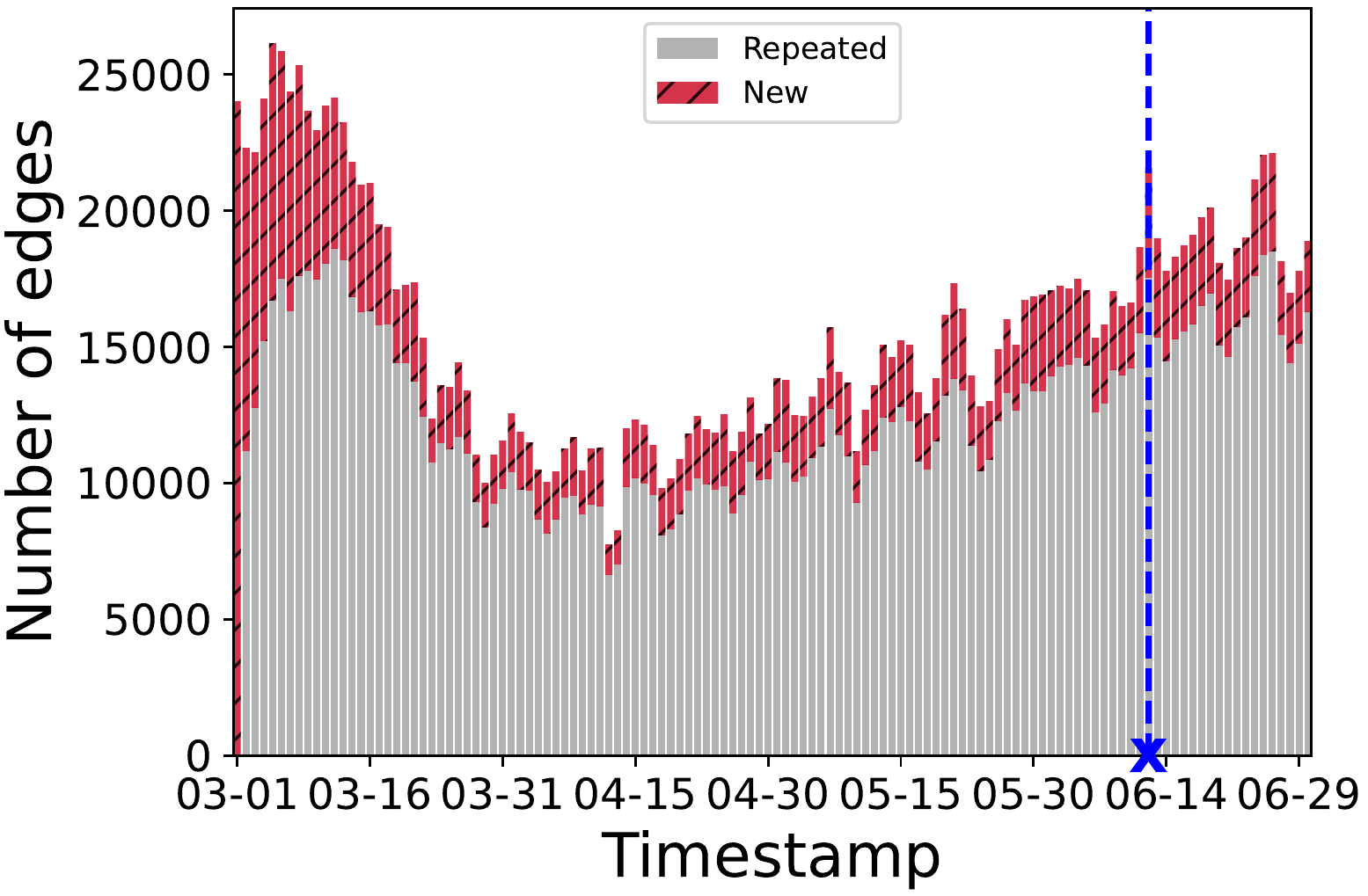}
  \caption{\Flights (0.21)}
\end{subfigure}%
\begin{subfigure}{0.25\textwidth}
 \includegraphics[width=\textwidth]{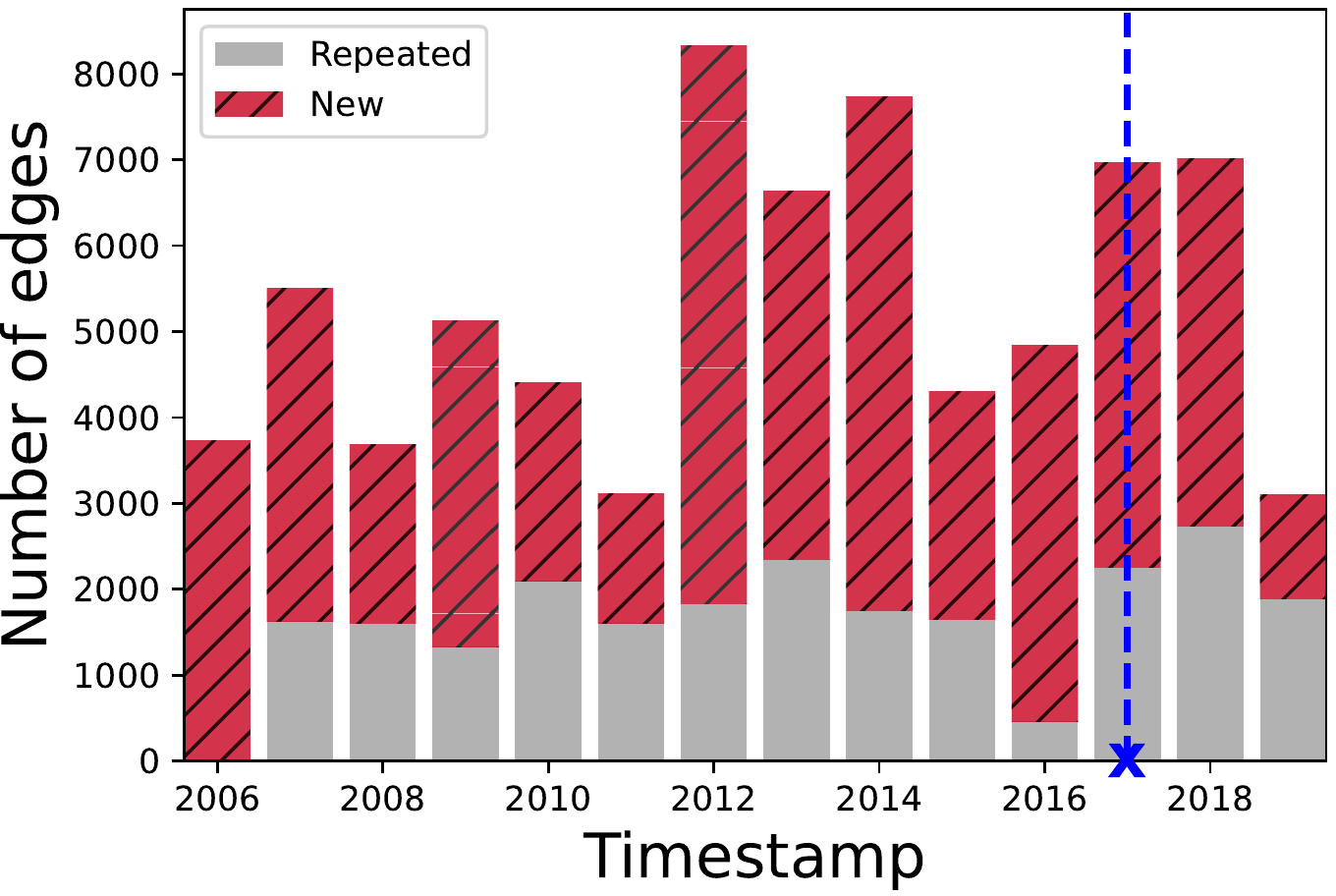}
  \caption{\CanParl (0.69)}
\end{subfigure}%

\begin{subfigure}{0.25\textwidth}
 \includegraphics[width=\textwidth]{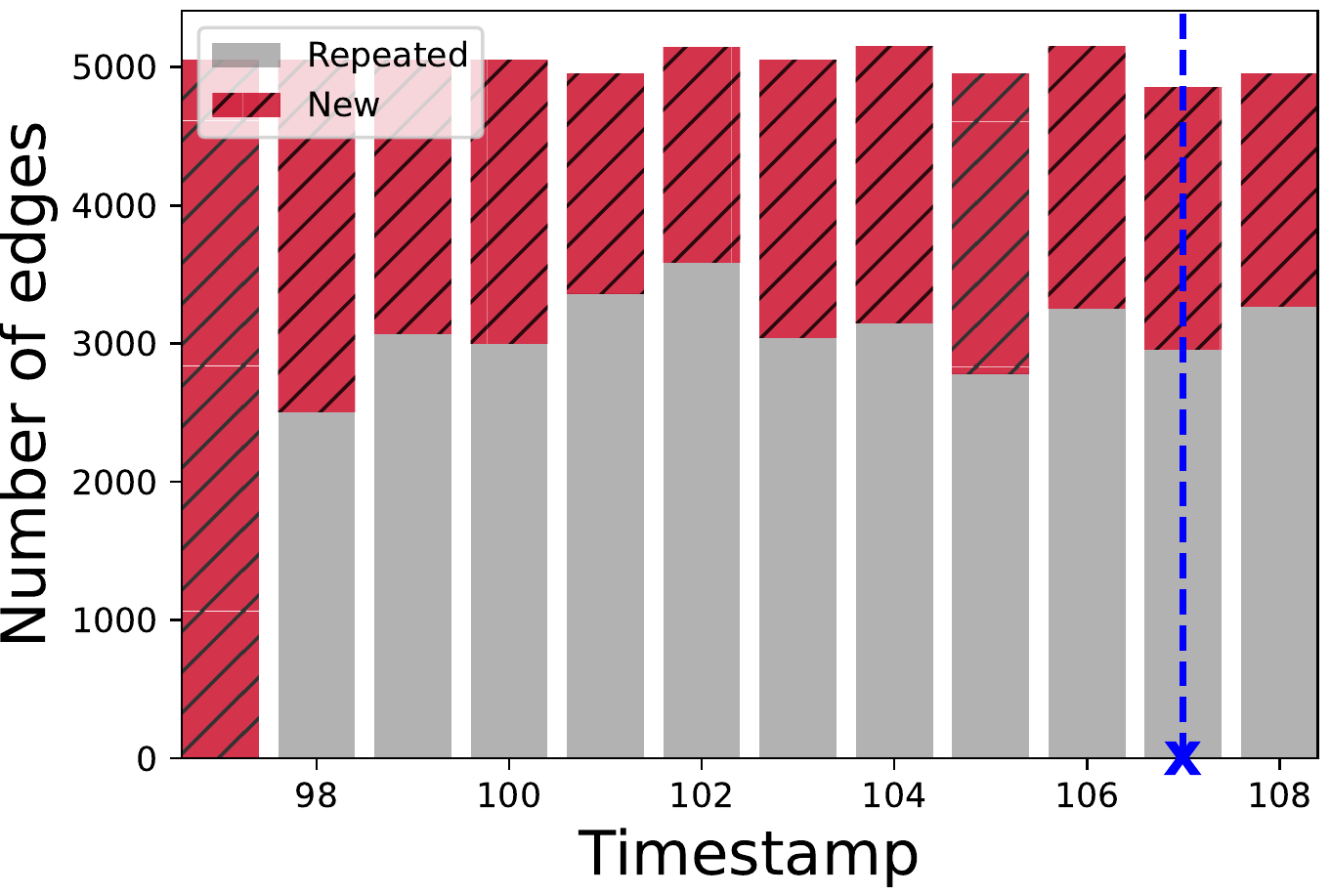}
  \caption{\USLegis (0.44)}
\end{subfigure}%
\begin{subfigure}{0.25\textwidth}
 \includegraphics[width=\textwidth]{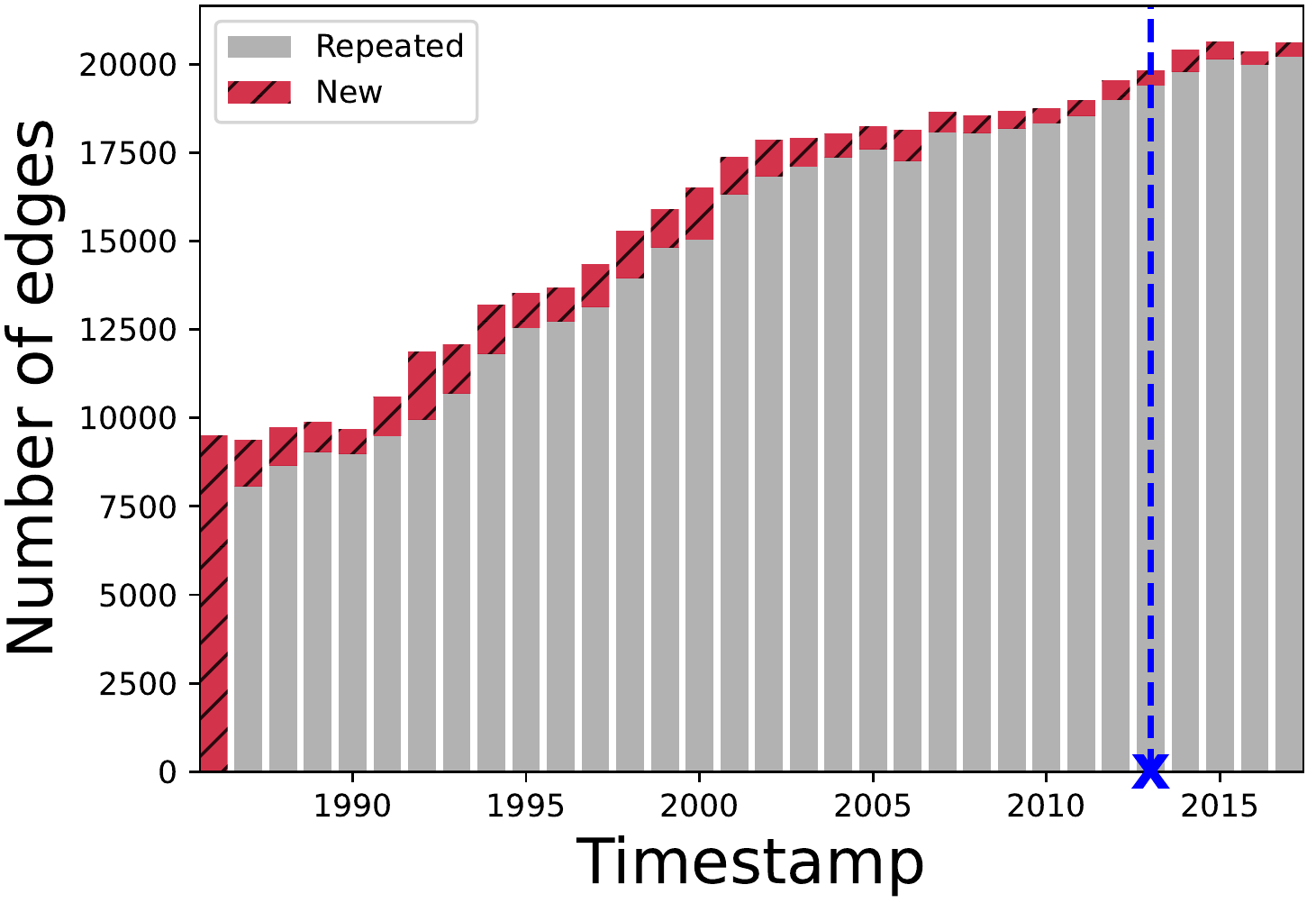}
  \caption{\UNTrade (0.07)}
\end{subfigure}%
\begin{subfigure}{0.25\textwidth}
 \includegraphics[width=\textwidth]{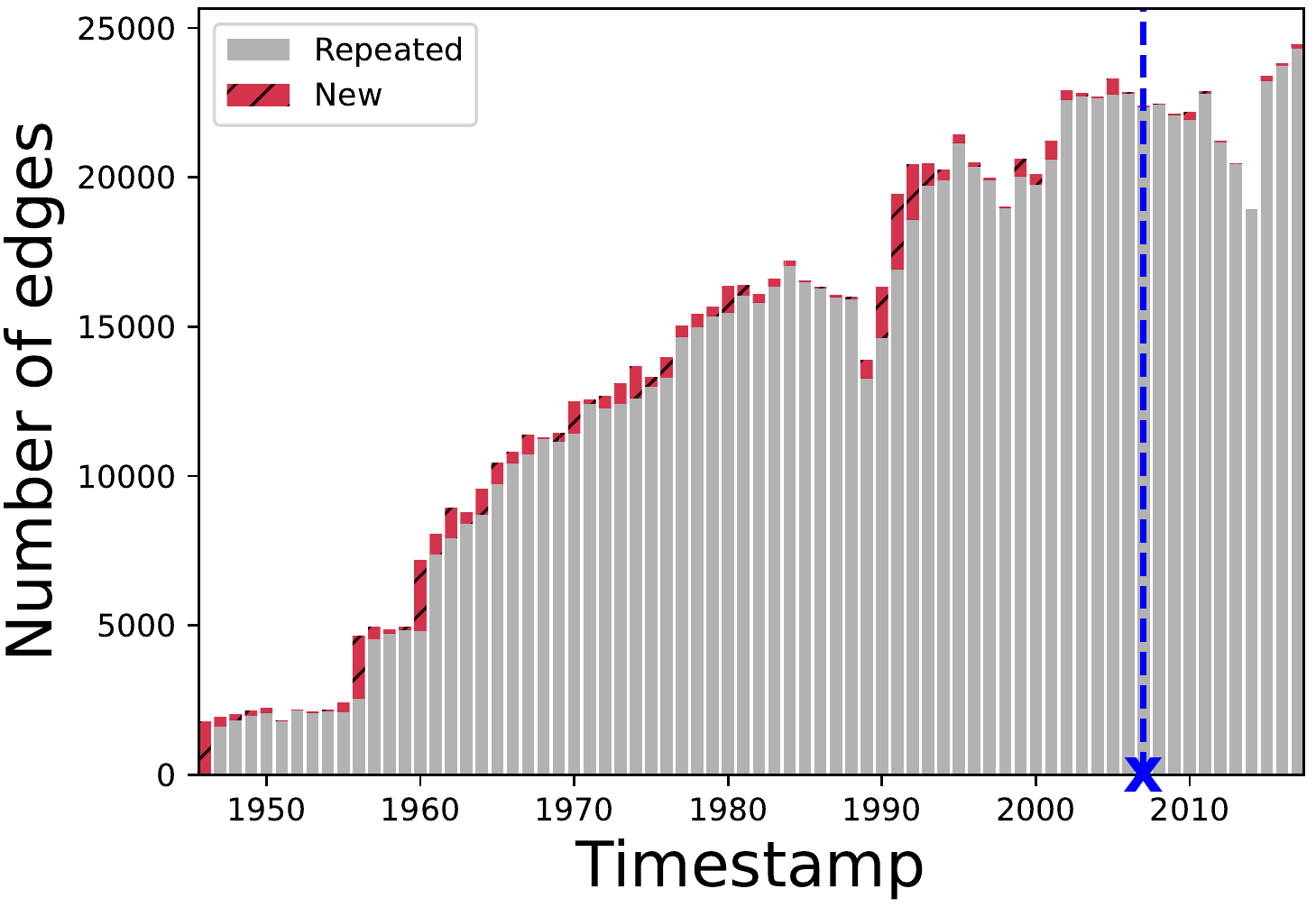}
  \caption{\UNVote (0.03)}
\end{subfigure}%
\begin{subfigure}{0.25\textwidth}
  \centering
  \includegraphics[width=\textwidth]{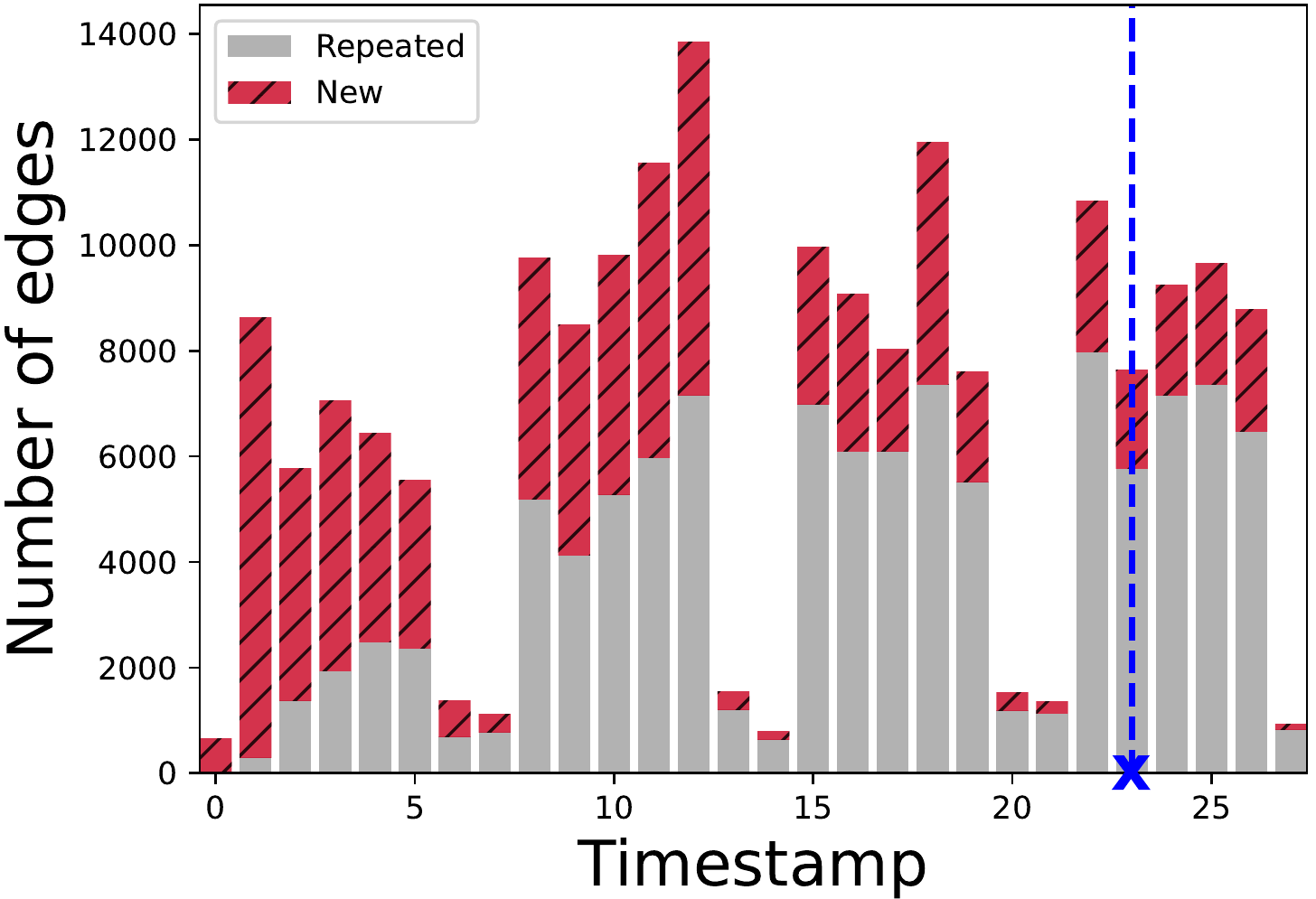}
  \caption{\changed{\contact (0.42)}}
\end{subfigure}%
\vspace{-5pt}
\caption{\TEA plots show many real-world dynamic networks contain a large proportion of edges that reoccur over time. 
Thus, even a simple memorization approach such as \method can potentially achieve strong performance.
The numbers in parentheses report the novelty index.
\changed{Due to space limitation, the Reddit's \TEA plot is presented in \figureRef{fig:reddit_TEA} in Appendix \ref{subsec:reddit_TEA_TET}.}
}
\label{fig:e_distribution}\vspace{-12pt}
\end{figure*}


\newcommand\negate[1]{\pgfmathsetmacro\result{1-#1}\result}

\begin{figure*}[t]\vspace{-12pt}
\centering
\begin{subfigure}{0.33\textwidth}
 \includegraphics[width=\textwidth]{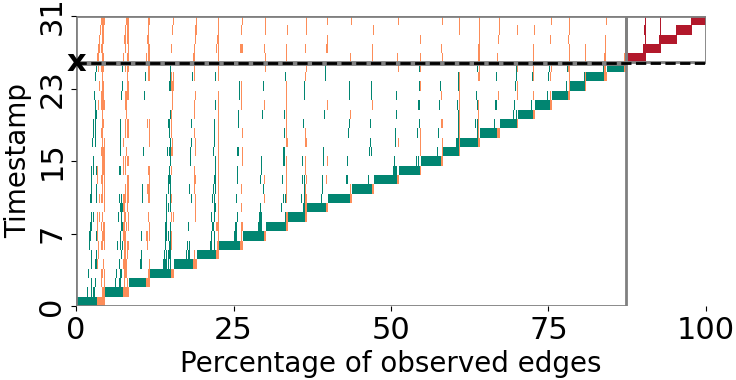}
  \caption{\Wikipedia (\negate{0.74} \& 0.42)}
\end{subfigure}%
\begin{subfigure}{0.33\textwidth}
  \centering
  \includegraphics[width=\columnwidth]{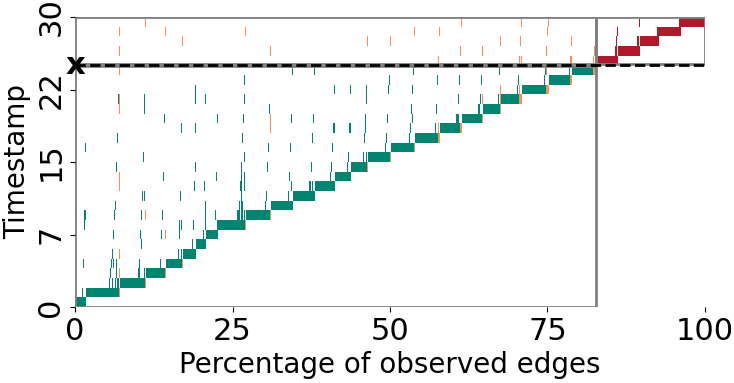}
  \caption{\MOOC (\negate{0.98} \& 0.79)}
\end{subfigure}%
\begin{subfigure}{0.33\textwidth}
  \centering
  \includegraphics[width=\columnwidth]{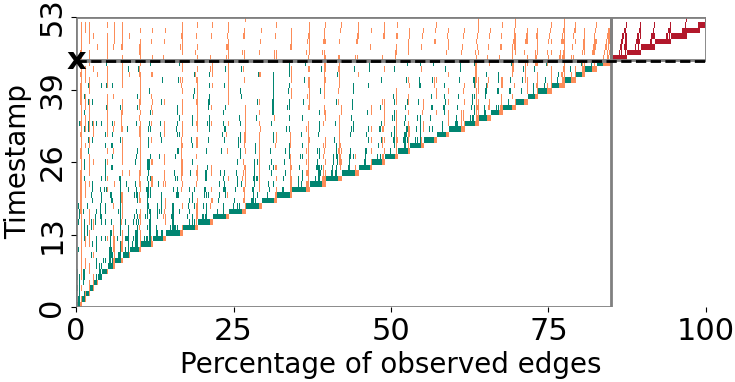}
  \caption{\LastFM (\negate{0.70} \& 0.35)}
\end{subfigure}%

\begin{subfigure}{0.33\textwidth}
 \includegraphics[width=\columnwidth]{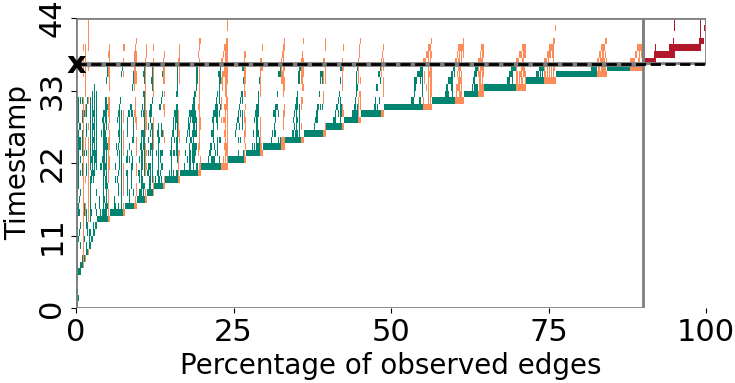}
  \caption{\Enron (\negate{0.78} \& 0.27)}
\end{subfigure}%
\begin{subfigure}{0.33\textwidth}
 \includegraphics[width=\columnwidth]{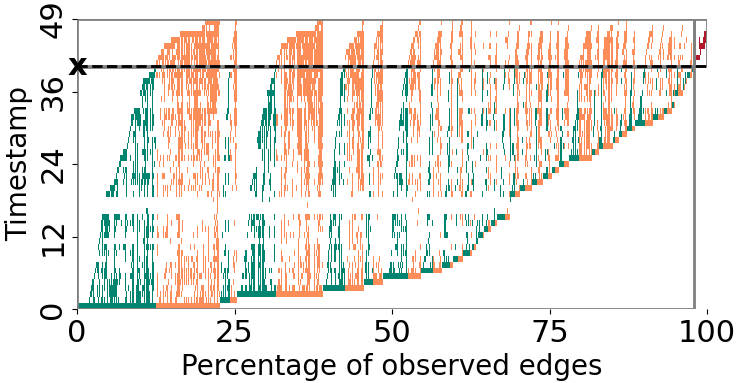}
  \caption{\SocialEvo (\negate{0.49} \& 0.02)}
  \label{fig:TET_SocialEvo}
\end{subfigure}%
\begin{subfigure}{0.33\textwidth}
  \centering
  \includegraphics[width=\columnwidth]{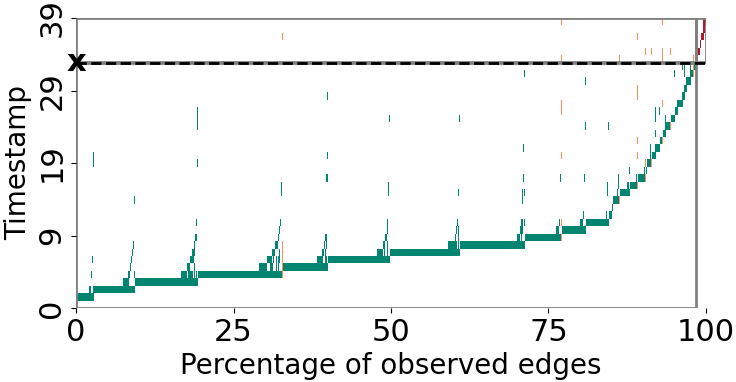}
  \caption{\UCI (\negate{0.99} \& 0.56)}
\end{subfigure}%

\begin{subfigure}{0.33\textwidth}
 \includegraphics[width=\columnwidth]{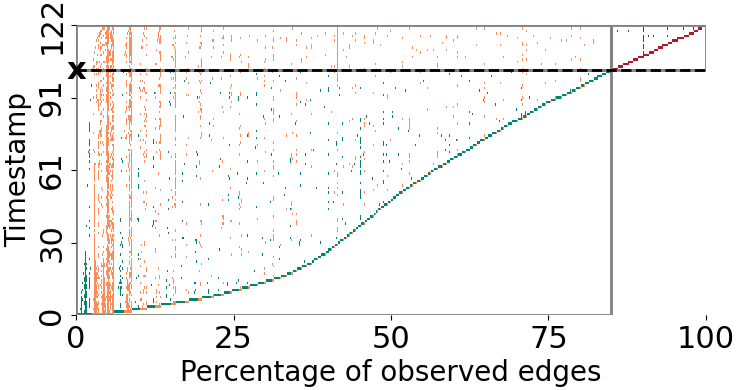}
  \caption{\Flights (\negate{0.40} \& 0.19)}
\end{subfigure}%
\begin{subfigure}{0.33\textwidth}
 \includegraphics[width=\columnwidth]{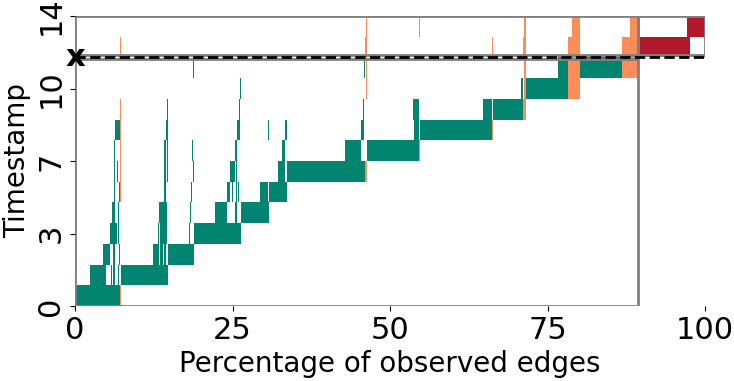}
  \caption{\CanParl (\negate{0.99} \& 0.57)}
\end{subfigure}%
\begin{subfigure}{0.33\textwidth}
  \centering
  \includegraphics[width=\columnwidth]{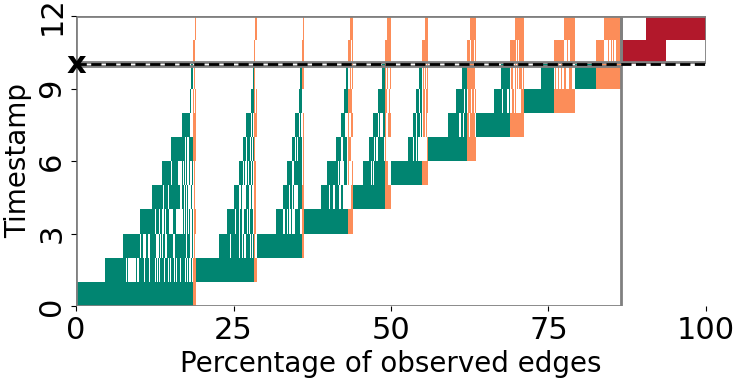} 
  \caption{\USLegis (\negate{0.92} \& 0.45)}
\end{subfigure}%

\begin{subfigure}{0.33\textwidth}
 \includegraphics[width=\columnwidth]{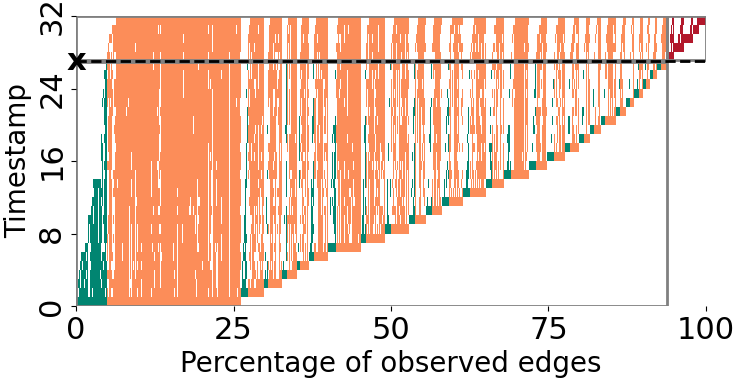}
  \caption{\UNTrade (\negate{0.13} \& 0.04)}
\end{subfigure}%
\begin{subfigure}{0.33\textwidth}
  \centering
  \includegraphics[width=\columnwidth]{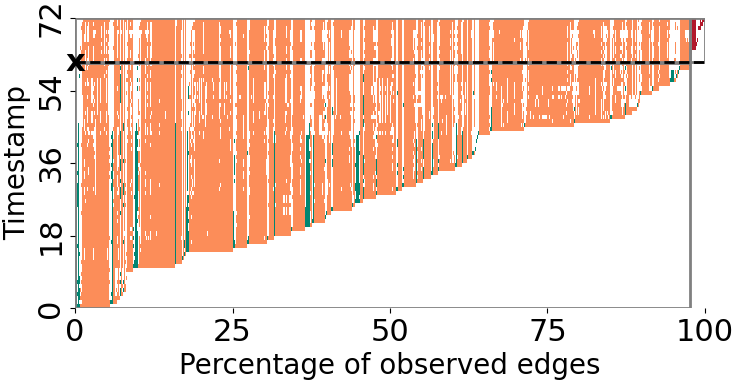}
  \caption{\UNVote (\negate{0.07} \& 0.01)}
\end{subfigure}%
\begin{subfigure}{0.33\textwidth}
  \centering
  \includegraphics[width=\columnwidth]{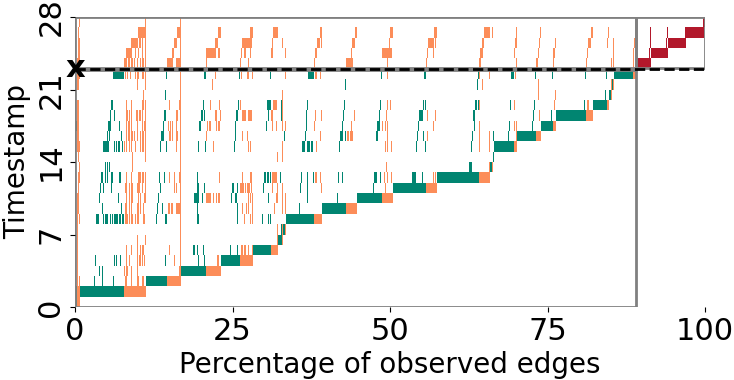}
  \caption{\changed{\contact (0.44 \& 0.12)}}
\end{subfigure}%
\vspace{2pt}
\begin{subfigure}{\textwidth}
\begin{center}
    \includegraphics[width=0.6\columnwidth]{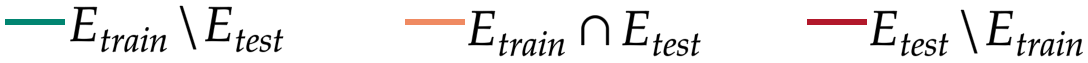}
\end{center}
\end{subfigure}%

\caption{\TET plots illustrates varied edge traffic patterns in different temporal graphs. 
The horizontal line starting with "\textbf{x}" marks $t_{split}$. 
In parentheses, we report the proportion of train edges reoccurring in the test set (reocurrence index) \& the proportion of unseen test edges (surprise index), respectively.
\changed{Due to space limitation, the Reddit's \TET plot is presented in \figureRef{fig:reddit_TET} in Appendix \ref{subsec:reddit_TEA_TET}.}
}
\label{fig:e_presence}\vspace{-14pt}
\end{figure*}

\subsection{Temporal Edge Appearance (\TEA) Plot}
\label{subsec:tea}

A \TEA plot illustrates the portion of repeated edges versus newly observed edges for each timestamp in a dynamic graph, as shown in \figureRef{fig:e_distribution}. 
The grey bar indicates the number of edges which were observed in previous time steps and the red bar represents the number of new edges seen at each step. To further quantify the observed pattern, we measure the average ratio of new edges in each timestamps as: 
\begin{align*} 
novelty = \frac{1}{T} \sum^T_{t=1} \frac{|E^t \setminus{E^t_{seen}}|}{|E^t|},\; \text{where} \; E^t = \{(s,d,t_e) | \; t_e = t \} \; \text{and}\; E^t_{seen} =\{(s,d,t_e) | \; t_e < t \}
\end{align*}
Here, $E^t$ denotes the set of edges present in timestamp $t$, and $E^t_{seen}$ denotes the set of all edges seen in the previous timestamps. This metric gives an estimation of the portion of the positive edges that a pure memorization method cannot predict correctly.

\figureRef{fig:e_distribution} shows high variance across datasets in temporal evolutionary patterns in terms of new and repeated edges. Some datasets such as \SocialEvo comprise mainly repeated edges, while others such as \MOOC have a high proportion of new edges.
The \TEA plots also show significant differences in when edges occur, and distinctions between our new datasets and existing ones. For example, our new \Flights dataset has significantly more unique edges and higher numbers of edges per timestamp. 

\TEA plots imply the importance of considering the relative distribution of the repeated and new edges when designing and choosing methods for the dynamic link prediction task. Because when many edges are repeated, a simple memorization approach can potentially achieve strong performance. On the other hand if there are many new edges, memorization cannot be sufficient. While the \TEA plot shows how many edges are repeated or new overall, it does not directly show how consistent the edge repeats are. Thus, we now propose:

\subsection{Temporal Edge Traffic (\TET) Plot}
\label{subsec:tet}

A \TET plot visualizes the reocurrence pattern of edges in different dynamic networks over time, as shown in \figureRef{fig:e_presence}. 
To construct these plots, we first sort edges based on the timestamp they first appear. Then for edges occurring in the same timestamp, we sort them based on when they last occur. Further, we color edges based on whether they are seen in train only (green), test only (inductive edges, red), or both (transductive edges, orange). To quantify the patterns in these plots we define the following two indices:
\begin{align*} 
{reocurrence} = 
\frac{|E_\text{train} \cap {E_\text{test}}|} {|E_\text{train}|}\; , \quad {surprise} = \frac{| E_\text{test}\setminus E_\text{train}|} {|E_\text{test}|}
\end{align*}




\TET plots provide more insights about the edges that are used for training and testing of different DGNN methods.
A memorization approach can potentially predict the transductive positive edges, since it has observed and hence recorded them during training. In particular, if they appear consistently, then simple memorization is likely to be successful. This is when reocurrence index is high and surprise index is low. On the other hand, if they appear at some times but then disappear later, then memory is likely still helpful, but simple and full memorization will not work. It would incorrectly predict that those edges still exist, i.e. when reocurrence index is low.
Meanwhile, memorization is not helpful at all for predicting inductive positive test edges at their first appearance, since these are new edges that have not been observed before, i.e. high surprise index.
%



We encourage researchers to investigate the proposed \TEA and \TET plots to get a more comprehensive overview of dynamic graphs in addition to the network statistics.
For example, while \SocialEvo and \UNTrade have a relatively similar proportion of repeated vs. new edges based on their \TEA plots, we see in their \TET plots that \UNTrade has far more consistent reocurrence. The clear difference we can observe in the visualization is mirrored in the results - the best model on \UNTrade is among the worst on \SocialEvo, and vice versa (\figureRef{fig:abs_perf}).

\section{\method: A Baseline for Dynamic Link Prediction }
\label{sec:edgebank}


We propose a pure memorization-based approach called \method, in order to understand whether memorizing past edges can be a competitive baseline. This is based on the observations above that many edges in dynamic graphs reoccur over time. 
The memory component of \method is simply a dictionary which is updated with newly observed edges at each timestamp, similar to the memory update procedure of TGN~\cite{rossi2020temporal}.
In this way, \method resembles a \textit{bank} of observed edges and requires no parameters. The storage requirement of \method is the same as the number of edges in the dataset.

At test time, \method predicts a test edge as \textit{positive} if the edge was seen before~(in the memory), and \textit{negative} otherwise. 
\method can predict correctly for edges which reoccur frequently over time. 
There are two scenarios where \method will make an incorrect prediction: (i) an unseen edge, or (ii) an edge observed before~(in memory) that is a negative edge at the current time. In the standard random negative sampling evaluation~\cite{rossi2020temporal,xu2020inductive,wang2020inductive}, as graphs are often sparse, it is unlikely that an edge observed before will be sampled as a negative edge. Therefore, \method has strong performance on negative edges in many cases. 

We consider two different memory update strategies for \method thus resulting in two variants:
\begin{itemize}[align=parleft,left=0pt..1em, noitemsep, topsep=0pt]
    \item $\textbf{\method}_\mathbf{\infty}$ stores all observed edges in memory, thus remembering edges even from a long time ago. 
    It is prone to false positives on edges which appear once but rarely reoccur over time. 
    \item $\textbf{\method}_\textbf{tw}$ only remembers edges from a fixed size time window from the immediate past. The size of the time window is set to the duration of test split, based on the intuition of predicting the test set behavior from the most similar (recent) period in the train set. 
    Hence, $\text{\method}_\text{tw}$ focuses on the edges observed in the short-term past. 
\end{itemize}


Note that \method is not designed to replace state-of-the-art methods. Rather we argue that all dynamic graph representation methods should be able to do better than memorization, thus beating \method. \method provides a simple and strong baseline to demonstrate how far pure memorization can go on each dataset.


\section{Revisiting Negative Sampling in Dynamic Graphs}
\label{sec:negative_sampling}

Current SOTA methods for dynamic link prediction often achieve near perfect performance on existing benchmark datasets \cite{kumar2019predicting, trivedi2019dyrep, xu2020inductive, rossi2020temporal, wang2020inductive, tian2021streaming}.
Consequently, one can argue that either the existing datasets are too simplistic or the current evaluation process is insufficient to differentiate methods. We discussed the dataset aspect extensively. 
Next, we need to carefully examine the current evaluation setting of DGNNs.
In particular, although negative edges constitute half of the evaluation edges, little attention has been dedicated to understanding the effect of different sets of negative edges on the overall performance. In this section, we take a closer look at Negative Sampling~(NS) strategies for evaluation of dynamic link prediction, and propose two novel NS strategies for more robust evaluation and better differentiation amongst methods. 
To better motivate the two new methods, we 
first explain the standard random NS strategy widely used in literature. 

\begin{wrapfigure}{R}{0.55\textwidth}\vspace{-12pt}
\centering
\includegraphics[width=0.55\textwidth]{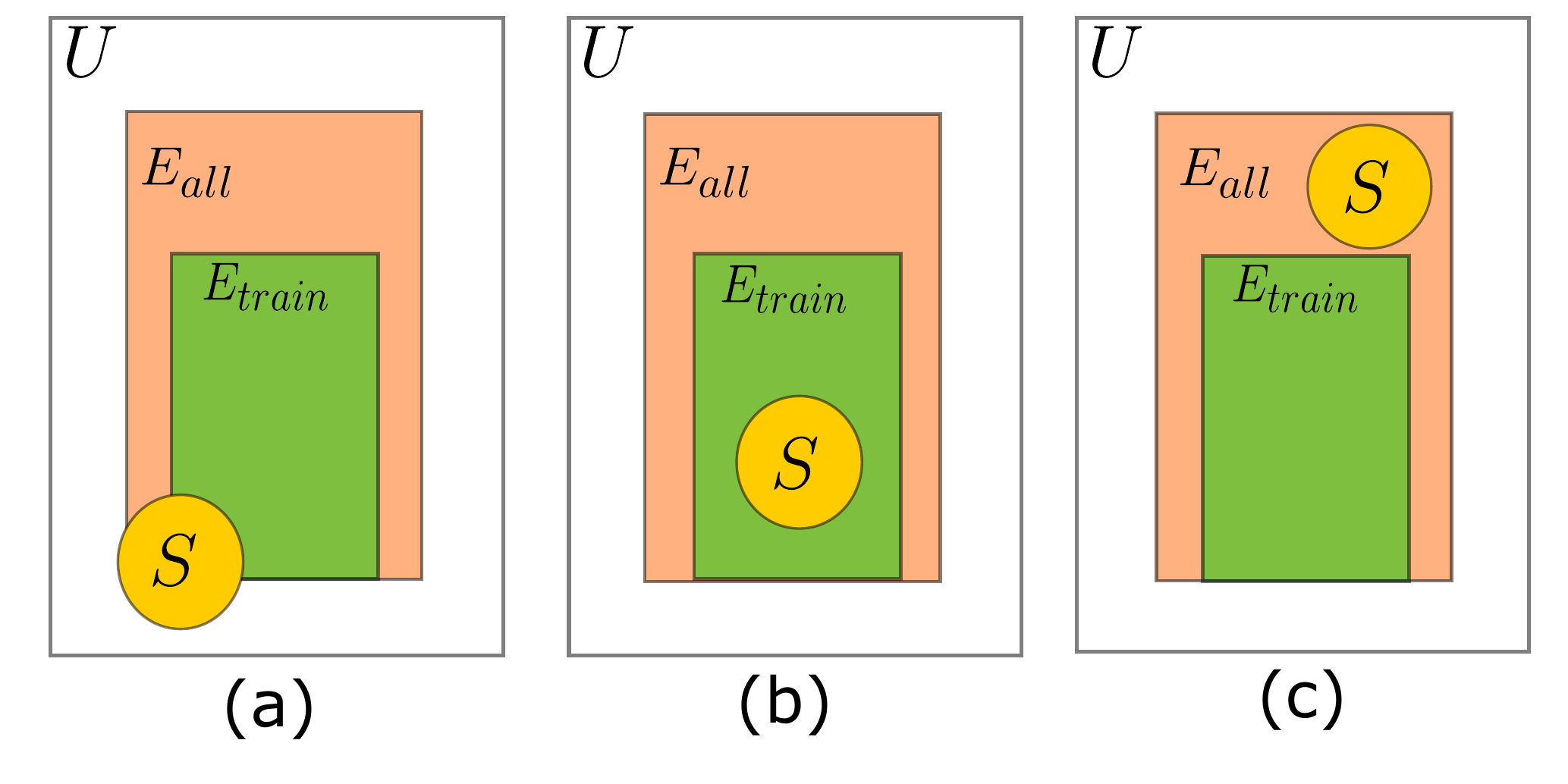}\vspace{-5pt}
\caption{Negative edge sampling strategies during evaluation for dynamic link prediction; (a) random sampling (standard in existing works), (b) historical sampling (ours), (c) inductive sampling (ours). }
\label{fig:vann}\vspace{-12pt}
\end{wrapfigure}

\textbf{Random Negative Sampling}
Current evaluation samples negative edges randomly from almost all possible node pairs of the graphs \cite{kumar2019predicting, trivedi2019dyrep, xu2020inductive, rossi2020temporal, wang2020inductive}.
At each time step, we have a set of positive edges consisting of source and destination nodes together with edge timestamps and edge features. To generate negative samples, the standard procedure is to keep the timestamps, features, and source nodes of the positive edges, while choosing destination nodes randomly from all nodes.
This approach has two significant issues:
    
     \textbf{(1) No Collision Checking}: \changed{most existing implementations have no collision check between positive and negative edges. There are some exceptions, such as  \cite{bielak2022fildne}, but this holds for all the DGNN methods examined in our experiments.} Therefore, it is possible for the same edge to be both positive and negative. This collision is more likely to happen in denser datasets, such as \UNVote and \UNTrade. A basic accept-reject sampling could address this issue, as applied  in our experiments. 
    
    \textbf{(2) No Reoccurring Edges}: the probability of sampling an edge which was observed before is often very low due to the sparsity of the graph. Therefore, a simple method like \method can perform well on negative edges. However, in many real-world tasks such as flight prediction, correct prediction of the same edge for different time steps is particularly important. For example, predicting that there will be no flight between the north and south poles this week is not nearly as practical as predicting whether a standard, reoccurring commuter flight will be canceled.
    
To address the second issue, we need to sample from previously observed edges, which can be from train or test set. This constitutes the two alternative NS strategies proposed here, illustrated in \figureRef{fig:vann}.
    %
%
Here, $S$ is the sample space for negative edges. 
Let $U$, $E_{all}$, $E_{train}$ be the set of all possible node pairs, all edges in the dataset (train and test) and all edges in the train set, respectively. Note that $E_{all} = E_{train} + E_{test}$ where $E_{test}$ is all edges in the test set. Lastly, we set $U_{neg} = U - E_{all}$. 
In random NS, we sample from edges $e \in U$, with the proportion from $E_{all}$ and $E_{train}$ regulated only by the sizes of those sets relative to U. To resolve the issues with random NS, in the following sections we propose \emph{historical NS} and \emph{inductive NS}. 

\textbf{Historical Negative Sampling.} \label{subsec:hist_ns}
In historical NS, we focus on sampling negative edges from the set of edges that have been observed during previous timestamps but are absent in the current step.
The objective of this strategy is to evaluate whether a given method is able to predict in which timestamps an edge would reoccur, rather than, for example, naively predicting it always reoccurs whenever it has been seen once. 
Therefore, in historical NS, for a given time step $t$, we sample from the edges $e \in ( E_{train} \cap  \overline{E_{t}})$.
Note that if the number of available historical edges is insufficient to match the number of positive edges, the remaining negative edges are sampled by the random NS strategy.


\textbf{Inductive Negative Sampling.} \label{subsec:induc_ns}
While in historical NS we focus on observed training edges, in inductive NS, our focus is to evaluate whether a given method can model the reocurrence pattern of edges only seen during test time. 
\changed{At test time, after observing the edges that were not seen during training, the model is asked to predict if such edges exist in future steps of the test phase.} 
Therefore, in the inductive NS, we sample from the edges $e \in (E_{test} \cap  \overline{E_{train}} \cap \overline{E_{t}})$ at time step $t$. As these edges are not observed during training, they are considered as \emph{inductive} edges.
Similar to before, if the number of inductive negative edges is not adequate, the remaining negative edges are sampled by the random NS strategy.

\begin{figure*}[t]\vspace{-12pt}
\centering
\begin{subfigure}{0.7\textwidth}
 \includegraphics[width=\textwidth]{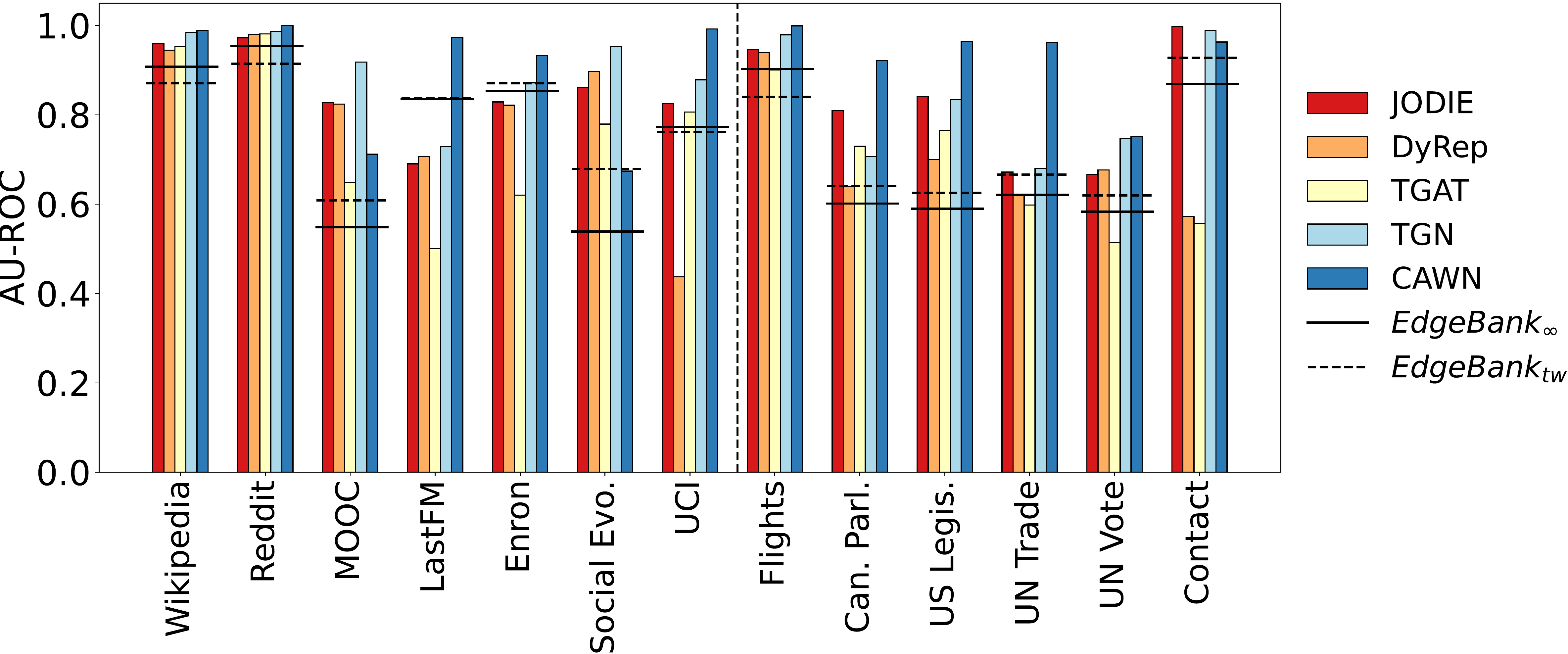}
  \caption{Standard random setting.}
  \label{fig:abs_perf_std}
\end{subfigure}%

\begin{subfigure}{0.5\textwidth}
 \includegraphics[width=\textwidth]{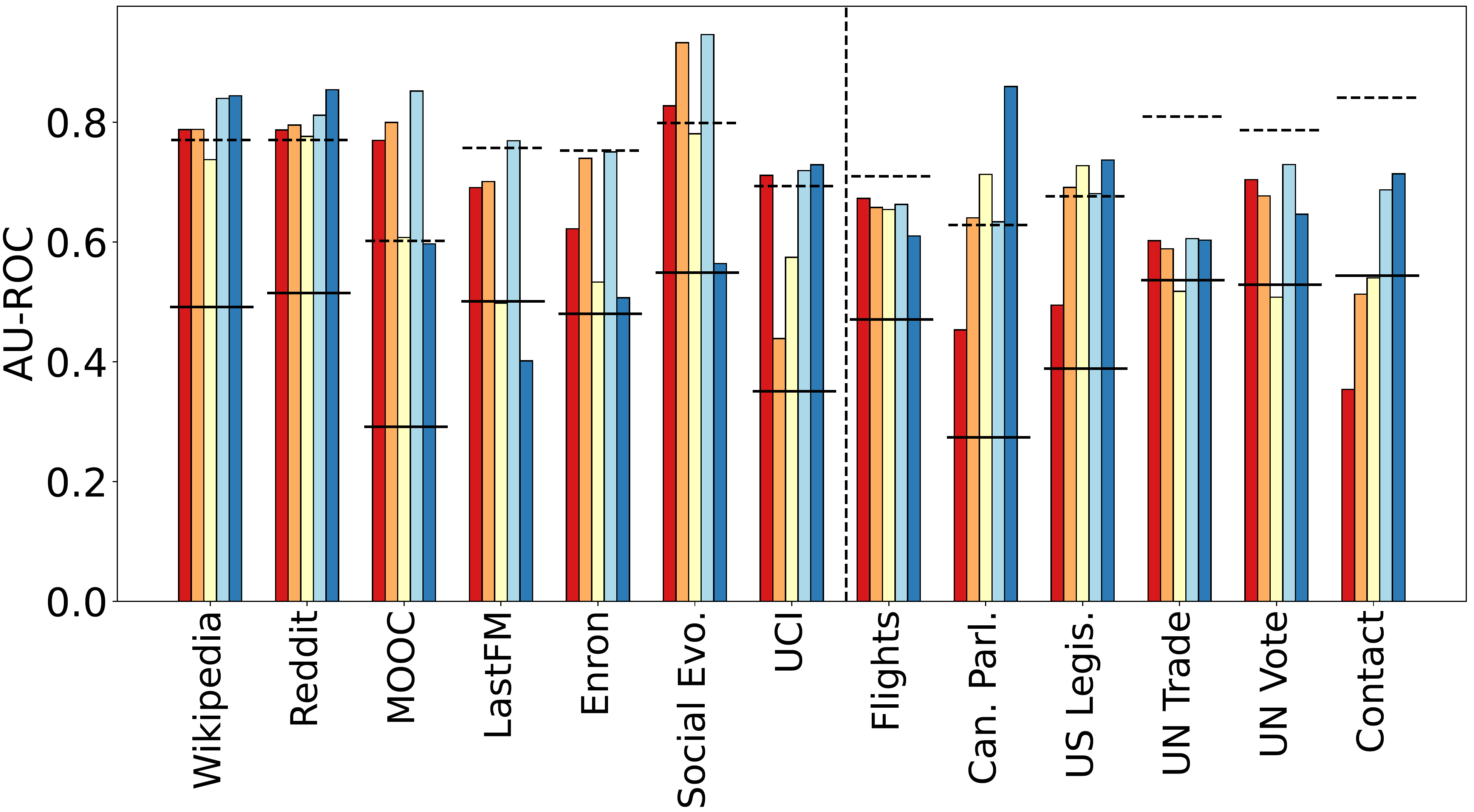}
  \caption{Historical setting.}
  \label{fig:abs_perf_hist}
\end{subfigure}%
\begin{subfigure}{0.5\textwidth}
 \includegraphics[width=\textwidth]{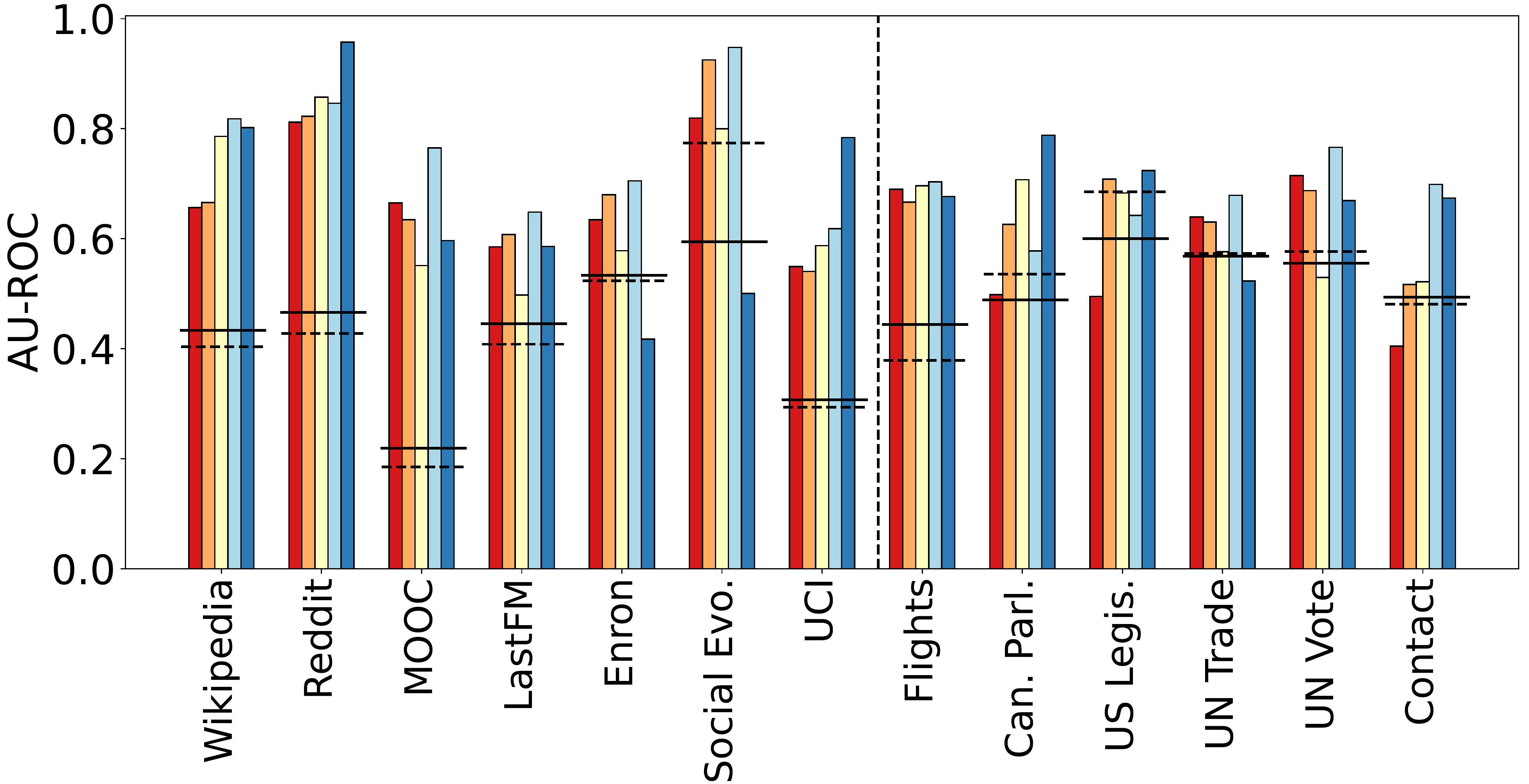}
  \caption{Inductive setting.}
  \label{fig:abs_perf_induc}
\end{subfigure}%

\begin{subfigure}{0.5\linewidth}
 \includegraphics[width=\linewidth]{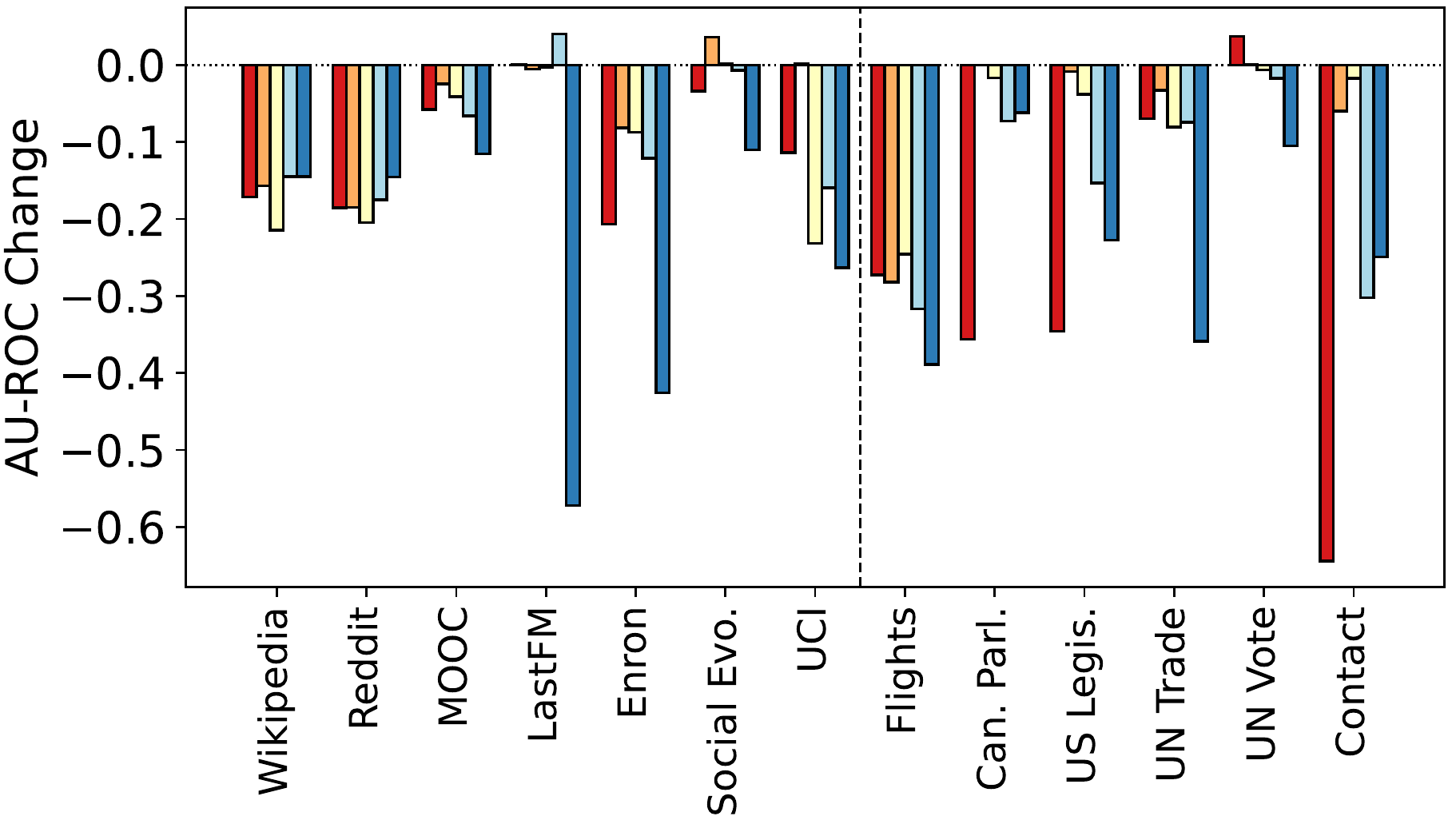}
  \caption{Historical compared to random NS.}
  \label{fig:delta_hist}
\end{subfigure}%
\begin{subfigure}{0.5\linewidth}
 \includegraphics[width=\linewidth]{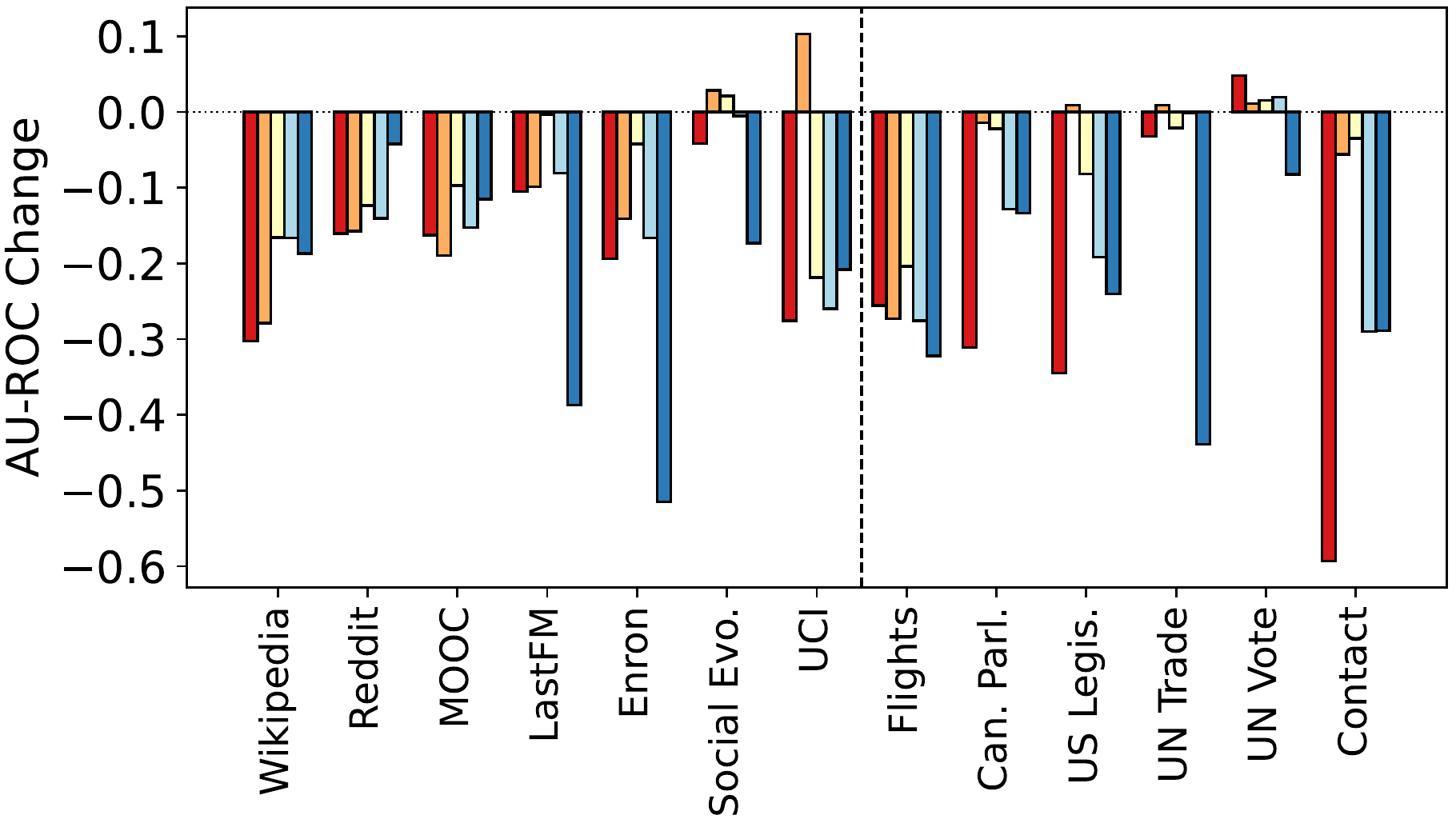}
  \caption{Inductive compared to random NS.}
  \label{fig:delta_induc}
\end{subfigure}%

\caption{Performance of methods in all three NS settings. In (a) the proposed memorization baselines are on par with SOTA methods, and over-performing in some datasets, e.g. LastFM. In (b) and (c), with alternative negative sampling strategies, we observe a more clear gap between the performance of models and the memorization baseline, whilst the ranking of the models also changes, e.g. CAWN not being the ranked one in most datasets, which is in contrast with the rankings obtained in the standard setting. In (d) and (e), we report the performance drop when moving from the standard setting, which can hint at the (lack of) generalization power of different methods, especially in (e). }
\vspace{-1em}
\label{fig:abs_perf}\vspace{-5pt}
\end{figure*}

\section{Experiments}\vspace{-5pt}
\label{sec:experiments}





In this section, we present a comprehensive evaluation of the dynamic link prediction task on all 13 datasets with 5 SOTA methods.
Our experimental setup closely follows~\cite{kumar2019predicting,trivedi2019dyrep,xu2020inductive,rossi2020temporal,wang2020inductive}. The objective of the link prediction task is to predict the existence of an edge between a node pair at a given time.  
For all DGNN methods, we use a Multilayer Perceptron as the output layer for edge prediction, where concatenated node embeddings are inputs and the probability of the edge is the output.
For all experiments, we use the same \textit{$70\%-15\%-15\%$} chronological splits for the train-validation-test sets as~\cite{xu2020inductive,rossi2020temporal,wang2020inductive}.
The averaged results over five runs are reported. The \textit{Area Under Receiver Operating Characteristic~(AU-ROC)} metric is selected as the main performance metric. We visualize the results for easier interpretation, but the exact numbers that produce the visualizations -- and the equivalents with \textit{Average Precision~(AP)} -- are presented in the Appendix~\ref{subsec:extended_results}.  

\begin{wrapfigure}{R}{0.5\textwidth}\vspace{-12pt}
\centering
\includegraphics[width=0.5\textwidth]{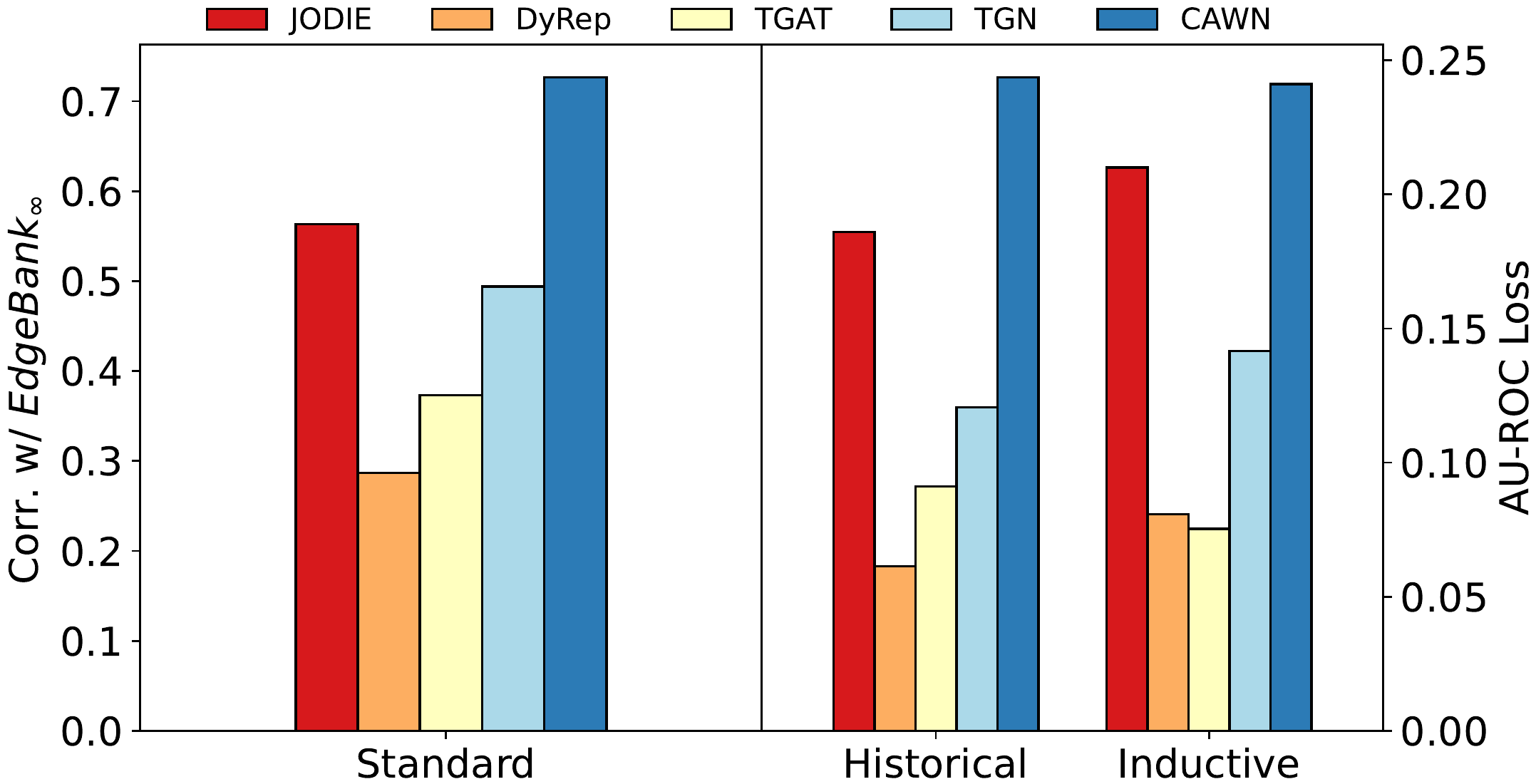}
\caption{Performance correlation with the proposed memorization baseline,  \textit{$\text{\method}_{\infty}$} (on the left), predicts the performance loss (lower = better) of the methods in both of the harder negative sampling settings (on the right).}
\label{fig:corr_eb}\vspace{-7pt}
\end{wrapfigure}





\figureRef{fig:abs_perf_std} compares the performance of all models under the standard random NS strategy. 
First, we observe significant variation in performance for all models across datasets. This supports the benefits of evaluation on datasets from different domains. Second, we observe a strong inconsistency in relative ranking amongst methods across datasets. For example, while CAWN achieves SOTA on most datasets, on \MOOC and \SocialEvo it performs significantly worse than several other models. Lastly, note that \method demonstrates competitive performance even when compared against SOTA methods. Despite being its simplicity, \method outperforms highly parametrized and complex models on datasets such as \LastFM, \Enron and \UNTrade.

Next, we examine the impact of NS strategies on performance. 
\figureRef{fig:abs_perf_hist} and \figureRef{fig:abs_perf_induc} shows the performance of different methods with the \emph{historical NS} and \emph{inductive NS} strategies, respectively. 
First, we observe that the ranking of models can change significantly across different NS settings. This shows that relying on a single NS strategy, such as the random NS, is insufficient for the complete evaluation of methods.
Second, for the historical NS setting, $\text{\method}_\text{tw}$ becomes highly competitive, often beating most methods and even achieving SOTA for \UNTrade, \UNVote, \Flights, \Enron, and \contact. This shows that in these datasets, recently observed edges contain crucial information for link prediction. Third, $\text{\method}_{\infty}$ has a significant drop in performance in both NS strategies. This shows that as the negative edges are sampled from either previously observed edges or unseen edges, naively memorizing all past edges is no longer sufficient. However, \method can perform competitively under random NS. This further shows that the standard random NS is limited in its ability to effectively differentiate methods. In \figureRef{fig:delta_hist} and \figureRef{fig:delta_induc}, we examine the performance changes for each model in historical or inductive NS setting. 
CAWN, which performed best overall with random NS, collapses on certain datasets such as \LastFM and \Enron. Other models fare much better on these datasets. All models exhibit a large performance drop on the \Flights dataset.

The performance degradation is also correlated with the degree of memorization. \figureRef{fig:corr_eb} shows that the models which are more correlated with $\text{\method}_{\infty}$ tend to perform worse in the historical and inductive NS settings. Since $\text{\method}_{\infty}$ is naively dependent on the memory, higher correlation with it indicates a model relies more heavily on memorization. For example, CAWN has the highest correlation and JODIE the second highest. They have the largest and second largest losses (respectively) in performance with the more challenging negative sampling. Similarly, DyRep is the least correlated with \method, and experiences the least drop in performance with historical NS and second least with inductive NS.

\section{Conclusion} \vspace{-7pt}
In this paper we proposed tools to improve evaluation of dynamic link prediction. First, we introduced \textit{six new datasets} to increase the diversity of domains in which link prediction methods are currently being evaluated. Then we created \TEA and \TET plots to visualize and quantify the temporal \textit{patterns} of edges in dynamic graphs, and the difficulty of an evaluation split. Next, we showed limitations of current random negative sampling strategy used in the evaluation and introduced two new strategies, \textit{historical} and \textit{inductive} sampling, to better test the generalization of different models. Finally, we proposed a competitive yet simple memorization-based \textit{baseline}, \method. It can yield insights into how much different models rely on memorization. When we applied these tools to compare existing models, we found that performance and ranking of different models vary significantly. We hope that these tools will lead to more thorough, lucid, and robust evaluation practices in dynamic link prediction.




\textbf{Acknowledgements:} This research is partially funded by the Canada CIFAR AI Chairs Program. The third author receives funding from IVADO. We thank Razieh Shirzadkhani for the help with cleaning and processing the \contact network.

\bibliographystyle{plainnat}
\bibliography{refs}


\appendix

\clearpage
 \onecolumn 

\section{Experiment Details}
\label{sec:appendix}

Here, we provide additional information regarding the baselines, datasets, and experimental settings.

\subsection{Baselines}
\label{subsec:baseline_methods}

We consider the following DGNN methods as our baselines.
\changed{All of these methods utilize node and edge features during the representation learning process if the network is attributed.
In case a network is not attributed, vectors of zeros are passed as the initial features.}

\begin{itemize}[align=parleft,left=0pt..1em,topsep=2pt]
\item \textbf{\href{https://github.com/srijankr/jodie}{JODIE}~\cite{kumar2019predicting}}: focuses on bipartite networks of instantaneous user-item interactions.
    JODIE has an update operation and a projection operation. The former utilizes two coupled RNNs to recursively update the representation of the users and items. 
    The latter predicts the future representation of a node, while considering the elapsed time since its last interaction. 
    \item \textbf{DyRep~\cite{trivedi2019dyrep}}: has a custom RNN that updates node representations upon observation of a new edge. 
    For obtaining the neighbor weights at each time, DyRep uses a temporal attention mechanism which is parameterized by the recurrent architecture. 
    \item \textbf{\href{https://github.com/StatsDLMathsRecomSys/Inductive-representation-learning-on-temporal-graphs}{TGAT}~\cite{xu2020inductive}}: aggregates features of temporal-topological neighborhood and temporal interactions of dynamic network. 
    The proposes TGAT layer employs a modified self-attention mechanism as its building block where the positional encoding module is replaced by a functional time encoding. 
    \item \textbf{\href{https://github.com/twitter-research/tgn}{TGN}~\cite{rossi2020temporal}}: 
    consists of five main modules: (1) \textit{memory}: containing each node's history and is used to store long-term dependencies, 
    (2) \textit{message function}: for updating the memory of each node based on the messages that are generated upon observation of an event, (3) \textit{message aggregator}: aggregating several message involving a single node, (4) \textit{memory updater}: responsible for updating the memory of a node according to the aggregated messages, and (5) \textit{embedding}: generating the representations of the nodes using the node's memory as well as the node and edge features. 
    Similar to TGAT, TGN also utilizes time encoding for effectively capturing the inter-event temporal information.
    \item \textbf{\href{https://github.com/snap-stanford/CAW}{CAWN}~\cite{wang2020inductive}}: 
    generates several Causal Anonymous Walks (CAWs) for each node, and uses these CAWs to generate relative node identities. 
    The identities together with the encoded elapsed time are used for encoding the CAWs by an RNN.
    Finally, the encodings of several CAWs are aggregated and fed to a MLP for predicting the probability of a link between two nodes.
\end{itemize}

The source code of baseline methods are publicly available under MIT licence or Apache License 2.0.

\subsection{Data Collection and Processing}
\label{subsec:collection}

\begin{itemize}[align=parleft,left=0pt..1em,topsep=2pt]
\item \href{https://zenodo.org/record/3974209/#.Yf62HepKguU}{\textbf{\Flights}~(\textit{new})~\cite{schafer2014bringing}}: \changed{is a directed dynamic flight network illustrating the development of the air traffic during the COVID-19 pandemic. This dataset is derived from the OpenSky dataset~\cite{strohmeier2021crowdsourced} and cleaned by Olive et al.~\footnote{\url{https://zenodo.org/record/3974209/\#.Yf62HepKguU}}~\cite{schafer2014bringing} with the traffic library~\cite{olive2019traffic}. It contains all flights observed by the OpenSky network's 2500 members since January 1st, 2020. To convert the dataset into a machine learning friendly format, we have: (1) removed all flights where either the source or the destination airport is missing, (2) aggregated the flights with shared origin and destination in the same day into weighted edges~(increment by 1 per flight), and (3) assigned a unique node ID to each airport. For future research, the final cleaned and processed edge-list is available. The edge weights specify the number of flights between two given airports in a day.}     
    

\item \href{https://github.com/shenyangHuang/LAD}{\textbf{\CanParl}~(\textit{new})~\cite{huang2020laplacian}}: \changed{the Canadian Parliament Political network dataset is originally collected and processed by Huang et al.~\cite{huang2020laplacian} for the network change point detection task. As political networks are currently unrepresented in existing benchmarks, we curate it for dynamic link prediction in this work. The dataset is collected from the \href{https://openparliament.ca/}{Open Parliament} initiative which documents the voting process inside the Canadian Parliament. There are 338 Members of Parliament~(MPs) where each one represents an electrical district who is elected for four years and can be re-elected. The network documents the collaborative efforts among MPs. Each bill has a sponsor MP and other MPs can vote positively or negatively towards this bill. A directed edge is added from a voter MP to the sponsor MP when the voter MP votes \textit{"yes"} for the sponsor MP's who sponsors the bill. The edge weights specify the number of times that the voter MP voted positively for the sponsor MP in a year. We assigned an unique node ID to each MP and anonymized the data.}
    
\item \href{https://github.com/shenyangHuang/LAD}{\textbf{\USLegis}~(\textit{new})~\cite{fowler2006legislative,huang2020laplacian}}: \changed{the US Legislative network documents social interactions between US legislators based on their co-sponsorship relations on bills presented during the 93rd-108th Congress. The dataset is originally collected by Fowler et al.~\cite{fowler2006legislative}, and then, processed into an edge-list format by Huang et al.~\cite{huang2020laplacian}. In the US House and Senate, each piece of legislative must be sponsored by a unique legislator. Other legislators can then choose to publicly express support for such bill by cosponsoring it. Therefore, \USLegis network falls in the same domain as \CanParl network where positive interactions between politicians are observed. The main difference being that \USLegis is undirected. If two congress persons cosponsor a bill, an undirected edge is formed between them. The edges are grouped into snapshots biannually~(i.e., per congress). Note that the edge weights specify the number of times two congress persons have cosponsored a bill in a given congress. We also assign a unique node ID to each congress person to anonymized the data.} 
    
    
\item \href{https://www.fao.org/faostat/en/#data/TM}{\textbf{\UNTrade}~(\textit{new})~\cite{macdonald2015rethinking}}: \changed{the United Nations food and agriculture trade dataset is originally collected, processed and disseminated by the Food and Agriculture Organization of the United Nations~\footnote{\url{https://www.fao.org/faostat/en/\#data/TM}}. The data is mainly provided by UNSD, Eurostat and other national authorities as needed. The trade data includes all food and agriculture products imported or exported annually by all the countries in the world. Please refer to~\cite{macdonald2015rethinking} for more details about the data. To process the data into a temporal graph format, we sum over the normalized import values across all food types of a country \textit{$v$} from another country \textit{$u$} as a weighted directed edge from \textit{$u$} to \textit{$v$}. Likewise, we sum over the normalized export values as weighted outgoing edges between two countries. The processed dataset contains annual data from 1986 to 2017, modelling trading relations between nations. The edge weights specify the total sum of normalized agriculture import or export values between two countries.}

    
\item \href{https://dataverse.harvard.edu/dataset.xhtml?persistentId=doi:10.7910/DVN/LEJUQZ}{\textbf{\UNVote}~(\textit{new})~\cite{LEJUQZ_2009,bailey2017estimating}}: \changed{is a dataset of roll-call votes in the UN General Assembly for 1946 to 2017. Each country in the United Nations can vote \textit{yes}, \textit{abstain}, \textit{no}, or \textit{absent} for a given UN bill~(such as amendments, security council elections, voting procedure and more). To convert the dataset into a temporal graph, we modelled collaborative votes between nations. For example, if a nation \textit{$u$} and another nation \textit{$v$} both voted \textit{yes} for a given bill, we add an undirected edge between them. In this way, the \UNVote network models the evolving political collaborations between nations at the United Nations. Note that the edge weights count the number of times two nations both voted \textit{yes} for a bill in a year.}
    
\item \href{https://springernature.figshare.com/articles/dataset/Metadata_record_for_Interaction_data_from_the_Copenhagen_Networks_Study/11283407/1}{\contact~(\textit{new})~\cite{sapiezynski2019interaction}}: \changed{the Copenhagen Contact Network describes the temporal evolution of the physical proximity of around 700 university students over a period of four weeks. This is achieved by collecting data from smartphones and estimating physical proximity via Bluetooth signal strength as part of the \textit{Copenhagen Networks Study}~\cite{sapiezynski2019interaction}. Considerable efforts are exercised to anonymize the data and preserve the participants' privacy as indicated in~\cite{sapiezynski2019interaction}. We cleaned the physical proximity network by removing users outside of the school as they are not tracked consistently and convert the dataset into a temporal edge-list format. Each participant is assigned a unique ID and edges between users indicate that they are within close proximity of each other. The proximity between participants is tracked every five minutes via Bluetooth and the edge weights indicate the Received Signal Strength~(RSSI) in units of dBm~(Decibel-miliwatt), with a smaller value indicating that two participants are closer in proximity.}
\end{itemize}

\subsection{\TEA and \TET plots for \Reddit Dataset}
\label{subsec:reddit_TEA_TET}
\begin{figure*}[t]
\centering
\begin{subfigure}{0.5\textwidth}
  \centering
  \includegraphics[width=0.7\textwidth]{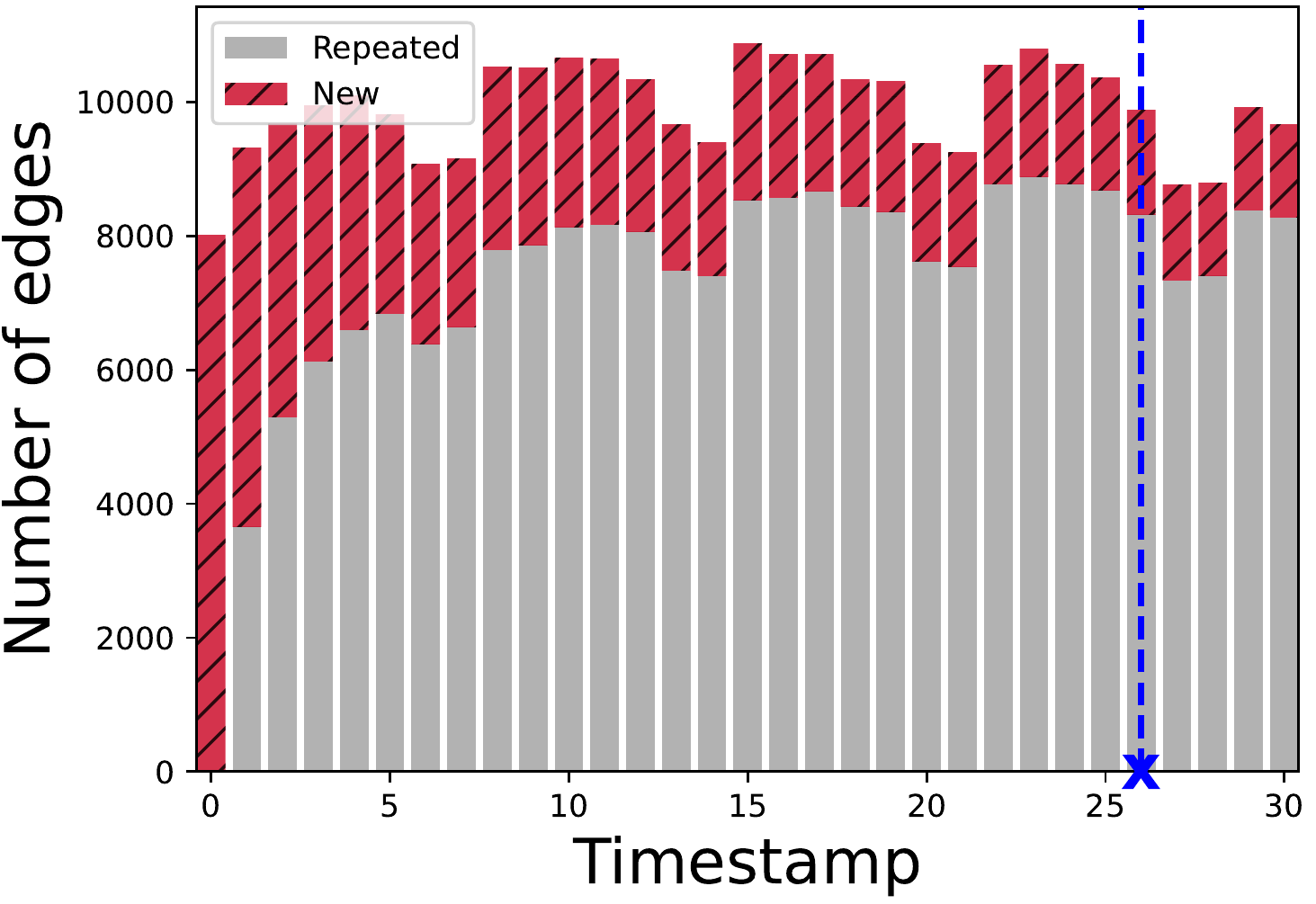}
  \caption{\TEA plot (0.26)}
  \label{fig:reddit_TEA}
\end{subfigure}%
\begin{subfigure}{0.5\textwidth}
  \centering
  \includegraphics[width=0.8\textwidth]{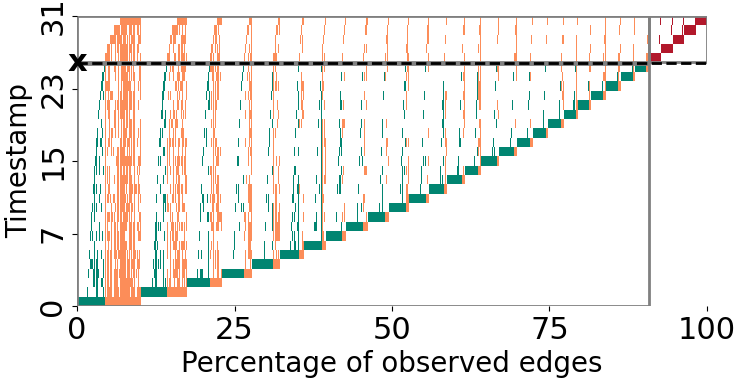}
  \caption{\TET plot (0.52 \& 0.18)}
  \label{fig:reddit_TET}
\end{subfigure}%
\caption{\TEA and \TET plots of \Reddit dataset.
The number in parentheses in (a) reports the novelty index.
In (b), we note the reoccurrence index \& surprise index in parentheses.
These indices are defined in Section \ref{sec:dataset}.
}
\label{fig:reddit_TEA_TET_plots}
\end{figure*}

\changed{Due to the space limitation in the main paper, we include the \TEA and \TET plots for the \Reddit dataset here in \figureRef{fig:reddit_TEA_TET_plots}.
}

\subsection{Discussion on Different Performance Metrics}
\label{subsec:diff_perf_metrics}
Essentially, the dynamic link prediction task is modeled as a binary classification problem by the existing literature.
Therefore, for evaluating the performance of different methods, utilization of threshold curve based metrics, such as Area Under the ROC cure (AU-ROC) and Average Precision (AP), are more encouraged \cite{yang2015evaluating}.
Other performance metrics (such as Accuracy, Precision, Recall, or \textit{$F_1$-score}) require a proper confidence threshold for their decision.
Since, the exact value of the threshold is often ill-defined in the literature, exploiting these metrics leads to unfair comparison across different methods.
In this work, we evaluated the performance of different methods in terms of AU-ROC (\tableRef{tab:au_roc_random_ns}, \tableRef{tab:au_roc_hist_ns}, and \tableRef{tab:au_roc_induc_ns}) and AP (\tableRef{tab:ap_random_ns}, \tableRef{tab:ap_hist_ns}, and \tableRef{tab:ap_induc_ns}).

\subsection{Hyperparameters} \label{app:hype}

For all the methods and datasets, we utilized the Adam optimizer with the learning rate equal to \textit{$0.0001$}.
The batch size for the training, validation, and testing was \textit{$200$}.
We set the number of epochs equal to \textit{$50$} and considered an early stopping with a patience of \textit{$5$}.
The dropout was \textit{$0.1$}, the number of attention heads was \textit{$2$}, and we set the node embedding size equal to \textit{$100$}.

For TGAT, TGN, and CAWN, the time embedding size was \textit{$100$}.
For TGN, we set the memory dimension as \textit{$172$} and message dimension as \textit{$100$}.
Lastly, for the CAWN, we used \textit{landing probability} as the positional encoding, and we set the positional embedding dimension as \textit{$64$} and the number of walk attention heads as \textit{$8$}.

\subsection{Computing Resources}
\label{subsec:computing_resources}

All the experiments were carried out on the \href{https://docs.alliancecan.ca/wiki/Graham}{\textit{Graham}} cluster of \textit{Compute Canada}. For all baselines, we executed each run with one of either P100, V100 or T4 Turing GPUs. The results reported are averaged over 5 runs and each run took on average 24 hours (depending on the dataset and the method). Therefore, each item reported in Table~\ref{tab:ap_random_ns} required 5 GPU days except for our proposed EdgeBank method which ran very efficiently on CPU only. 

\section{Additional Experimental Results}
\label{sec:additional_experimental_results}

\subsection{Extended Results}
\label{subsec:extended_results}
Here, we report the extended results used to plot the figures in the main paper. 
In particular, we report the AP and AU-ROC of the dynamic link prediction task in \textit{random} (\tableRef{tab:ap_random_ns} and \tableRef{tab:au_roc_random_ns}), \textit{historical} (\tableRef{tab:ap_hist_ns} and \tableRef{tab:au_roc_hist_ns}), or \textit{inductive} (\tableRef{tab:ap_induc_ns} and \tableRef{tab:au_roc_induc_ns}) NS setting.
We also report the AP loss and AU-ROC loss of \textit{historical} (\tableRef{tab:ap_hist_ns_difference} and \tableRef{tab:au_roc_hist_ns_difference}) and \textit{inductive} (\tableRef{tab:ap_induc_ns_difference} and \tableRef{tab:au_roc_induc_ns_difference}) compared to \textit{random} NS setting.

\begin{table*}[ht]
\caption{AP of dynamic link prediction in standard setting with \textit{random} negative sampling. The results report the mean over five runs with the standard deviations in parenthesis.}
\label{tab:ap_random_ns}
\resizebox{\linewidth}{!}{%
\begin{tabular}{l c c c c c | c c }
	\toprule \toprule
		Data & JODIE & DyRep & TGAT & TGN & CAWN & $\text{\method}_\text{tw}$ & $\text{\method}_\infty$ \\ 
	\midrule
		\Wikipedia & 0.95 (0.005) & 0.95 (0.004) & 0.95 (0.002) & 0.99 (0.001) & 0.99 (0.003) & 0.87 (0.000) & 0.90 (0.000) \\
		\Reddit & 0.95 (0.004) & 0.98 (0.001) & 0.98 (0.000) & 0.99 (0.000) & 0.99 (0.000) & 0.91 (0.000) & 0.95 (0.000) \\ 
		\MOOC & 0.78 (0.023) & 0.80 (0.016) & 0.61 (0.016) & 0.90 (0.010) & 0.75 (0.024) & 0.58 (0.000) & 0.53 (0.000) \\
		\LastFM & 0.68 (0.016) & 0.71 (0.015) & 0.50 (0.004) & 0.72 (0.055) & 0.98 (0.002) & 0.79 (0.000) & 0.77 (0.000) \\
		\Enron & 0.78 (0.023) & 0.80 (0.029) & 0.59 (0.024) & 0.85 (0.025) & 0.95 (0.003) & 0.84 (0.000) & 0.80 (0.000) \\ 
		\SocialEvo & 0.79 (0.063) & 0.87 (0.006) & 0.76 (0.007) & 0.93 (0.002) & 0.72 (0.189) & 0.61 (0.000) & 0.52 (0.000) \\
		\UCI & 0.75 (0.021) & 0.46 (0.072) & 0.78 (0.007) & 0.88 (0.021) & 0.99 (0.001) & 0.76 (0.000) & 0.76 (0.000) \\ 
	\midrule
	    \Flights & 0.94 (0.018) & 0.93 (0.007) & 0.89 (0.003) & 0.98 (0.003) & 0.99 (0.001) & 0.84 (0.000) & 0.89 (0.000) \\
	    \CanParl & 0.75 (0.014) & 0.58 (0.021) & 0.68 (0.033) & 0.64 (0.54) & 0.94 (0.036) & 0.65 (0.000) & 0.60 (0.000) \\ 
	    \USLegis & 0.76 (0.067) & 0.64 (0.076) & 0.70 (0.013) & 0.77 (0.037) & 0.97 (0.021) & 0.58 (0.000) & 0.55 (0.000) \\ 
	    \UNTrade & 0.64 (0.006) & 0.61 (0.007) & 0.58 (0.044) & 0.64 (0.016) & 0.97 (0.007) & 0.60 (0.000) & 0.57 (0.000) \\ 
	    \UNVote & 0.64 (0.011) & 0.64 (0.004) & 0.52 (0.002) & 0.71 (0.016) & 0.82 (0.022) & 0.57 (0.000) & 0.55 (0.000) \\ 
	    \contact & 0.99 (0.000) & 0.56 (0.051) & 0.58 (0.007) & 0.99 (0.003) & 0.97 (0.001) & 0.89 (0.000) & 0.80 (0.000) \\
	\bottomrule
	\bottomrule
	\end{tabular}
	}
\end{table*}

\begin{table*}[ht]
\caption{AU-ROC of dynamic link prediction in standard setting with \textit{random} negative sampling. The results report the mean over five runs with the standard deviations in parenthesis.}
\label{tab:au_roc_random_ns}
\resizebox{\linewidth}{!}{%
\begin{tabular}{l c c c c c | c c }
	\toprule \toprule
		Data & JODIE & DyRep & TGAT & TGN & CAWN & $\text{\method}_\text{tw}$ & $\text{\method}_\infty$ \\ 
	\midrule
		\Wikipedia & 0.96 (0.003) & 0.94 (0.004) & 0.95 (0.002) & 0.98 (0.001) & 0.99 (0.005) & 0.87 (0.000) & 0.91 (0.000) \\ 
		\Reddit & 0.97 (0.002) & 0.98 (0.001) & 0.98 (0.000) & 0.99 (0.000) & 0.99 (0.000) & 0.91 (0.000) & 0.95 (0.000) \\ 
		\MOOC & 0.83 (0.017) & 0.82 (0.014) & 0.65 (0.021) & 0.91 (0.010) & 0.71 (0.040) & 0.61 (0.000) & 0.55 (0.000) \\
		\LastFM & 0.69 (0.011) & 0.71 (0.009) & 0.50 (0.001) & 0.73 (0.049) & 0.97 (0.003) & 0.84 (0.000) & 0.84 (0.000) \\
		\Enron & 0.83 (0.017) & 0.82 (0.024) & 0.62 (0.021) & 0.87 (0.028) & 0.93 (0.003) & 0.87 (0.000) & 0.85 (0.000) \\ 
		\SocialEvo & 0.86 (0.042) & 0.90 (0.004) & 0.78 (0.007) & 0.95 (0.001) & 0.67 (0.195) & 0.68 (0.000) & 0.54 (0.000) \\ 
		\UCI & 0.83 (0.016) & 0.44 (0.103) & 0.81 (0.006) & 0.88 (0.020) & 0.99 (0.002) & 0.76 (0.000) & 0.77 (0.000) \\ 
	\midrule
	    \Flights & 0.95 (0.014) & 0.94 (0.006) & 0.90 (0.003) & 0.80 (0.002) & 0.99 (0.001) & 0.84 (0.000) & 0.90 (0.000) \\
	    \CanParl & 0.81 (0.011) & 0.64 (0.033) & 0.73 (0.040) & 0.71 (0.089) & 0.92 (0.050) & 0.64 (0.000) & 0.60 (0.000) \\ 
	    \USLegis & 0.84 (0.048) & 0.70 (0.092) & 0.77 (0.015) & 0.83 (0.033) & 0.96 (0.030) & 0.63 (0.000) & 0.59 (0.000) \\ 
	    \UNTrade & 0.67 (0.004) & 0.62 (0.023) & 0.60 (0.054) & 0.68 (0.015) & 0.96 (0.011) & 0.67 (0.000) & 0.62 (0.000) \\ 
	    \UNVote & 0.67 (0.010) & 0.68 (0.004) & 0.51 (0.001) & 0.75 (0.016) & 0.75 (0.026) & 0.62 (0.000) & 0.58 (0.000) \\ 
	    \contact & 0.99 (0.000) & 0.57 (0.051) & 0.56 (0.006) & 0.99 (0.002) & 0.96 (0.001) & 0.93 (0.000) & 0.87 (0.000)\\
	\bottomrule
	\bottomrule
	\end{tabular}
	}
\end{table*}

\begin{table*}[ht]
\caption{AP of dynamic link prediction in \textit{historical} negative sampling setting. The results report the mean over five runs with the standard deviations in parenthesis.}
\label{tab:ap_hist_ns}
\resizebox{\linewidth}{!}{%
\begin{tabular}{l c c c c c | c c }
	\toprule \toprule
		Data & JODIE & DyRep & TGAT & TGN & CAWN & $\text{\method}_\text{tw}$ & $\text{\method}_\infty$ \\ 
	\midrule
		\Wikipedia & 0.77 (0.014) & 0.81 (0.005) & 0.76 (0.007) & 0.88 (0.003) & 0.89 (0.048) & 0.71 (0.001) & 0.50 (0.000) \\ 
		\Reddit & 0.77 (0.008) & 0.79 (0.003) & 0.77 (0.004) & 0.81 (0.002) & 0.89 (0.022) & 0.70 (0.001) & 0.51 (0.000) \\ 
		\MOOC & 0.70 (0.058) & 0.74 (0.016) & 0.59 (0.033) & 0.84 (0.019) & 0.66 (0.278) & 0.57 (0.001) & 0.43 (0.000) \\
		\LastFM & 0.68 (0.067) & 0.71 (0.031) & 0.50 (0.001) & 0.76 (0.067) & 0.56 (0.070) & 0.69 (0.000) & 0.50 (0.000) \\
		\Enron & 0.56 (0.013) & 0.71 (0.012) & 0.53 (0.022) & 0.72 (0.30) & 0.63 (0.092) & 0.68 (0.002) & 0.50 (0.000) \\ 
		\SocialEvo & 0.73 (0.055) & 0.93 (0.004) & 0.77 (0.009) & 0.95 (0.005) & 0.64 (0.107) & 0.71 (0.000) & 0.53 (0.000) \\ 
		\UCI & 0.62 (0.089) & 0.45 (0.061) & 0.61 (0.008) & 0.76 (0.019) & 0.79 (0.070) & 0.65 (0.001) & 0.44 (0.001) \\ 
	\midrule
	    \Flights & 0.65 (0.014) & 0.63 (0.009) & 0.65 (0.004) & 0.64 (0.011) & 0.63 (0.067) & 0.65 (0.000) & 0.49 (0.000) \\
	    \CanParl & 0.43 (0.014) & 0.57 (0.015) & 0.67 (0.40) & 0.56 (0.044) & 0.90 (0.017) & 0.64 (0.001) & 0.48 (0.000) \\ 
	    \USLegis & 0.45 (0.025) & 0.63 (0.75) & 0.63 (0.062) & 0.56 (0.29) & 0.82 (0.129) & 0.63 (0.003) & 0.46 (0.001) \\ 
	    \UNTrade & 0.56 (0.012) & 0.58 (0.004) & 0.51 (0.009) & 0.57 (0.012) & 0.72 (0.015) & 0.73 (0.001) & 0.52 (0.000) \\ 
	    \UNVote & 0.66 (0.021) & 0.64 (0.007) & 0.51 (0.005) & 0.67 (0.019) & 0.75 (0.027) & 0.71 (0.001) & 0.51 (0.000) \\ 
	    \contact & 0.40 (0.001) & 0.51 (0.032) & 0.57 (0.009) & 0.57 (0.032) & 0.80 (0.013) & 0.77 (0.000) & 0.52 (0.000) \\
	\bottomrule
	\bottomrule
	\end{tabular}
	}
\end{table*}

\begin{table*}[ht]
\centering
\caption{The AP loss of \textit{historical} compared to random negative sampling. The intensity of the color relates to the amount of loss.}
\label{tab:ap_hist_ns_difference}
\begin{tabular}{l c c c c c | c c }
	\toprule \toprule
		Data & JODIE & DyRep & TGAT & TGN & CAWN & $\text{\method}_\text{tw}$ & $\text{\method}_\infty$ \\ 
	\midrule
		\Wikipedia & \cellcolor{q1}0.18 & \cellcolor{q1}0.14 & \cellcolor{q1}0.19 & \cellcolor{q1}0.11 & \cellcolor{q1}0.10 & \cellcolor{q1}0.16 & \cellcolor{q3}0.41 \\ 
		
		\Reddit & \cellcolor{q1}0.18 & \cellcolor{q1}10.19 & \cellcolor{q2}0.21 & \cellcolor{q1}0.18 & \cellcolor{q1}0.11 & \cellcolor{q2}0.21 & \cellcolor{q3}0.44 \\ 
		
		\MOOC & \cellcolor{q1}0.08 & \cellcolor{q1}0.06 & \cellcolor{q1}0.02 & \cellcolor{q1}0.06 & \cellcolor{q1}0.09 & \cellcolor{q1}0.01 & \cellcolor{q1}0.10 \\
		
		\LastFM & \cellcolor{q1}0.00 & \cellcolor{q1}0.00 & \cellcolor{q1}0.00 & \cellcolor{q1}-0.05 & \cellcolor{q3}0.42 & \cellcolor{q1}0.10 & \cellcolor{q2}0.27 \\
		
		\Enron & \cellcolor{q2}0.22 & \cellcolor{q1}0.08 & \cellcolor{q1}0.06 & \cellcolor{q1}0.13 & \cellcolor{q3}0.32 & \cellcolor{q1}0.15 & \cellcolor{q3}0.31 \\ 
		
		\SocialEvo & \cellcolor{q1}0.07 & \cellcolor{q1}-0.06 & \cellcolor{q1}-0.01 & \cellcolor{q1}-0.01 & \cellcolor{q1}0.08 & \cellcolor{q1}-0.10 & \cellcolor{q1}-0.01 \\ 
		
		\UCI & \cellcolor{q1}0.13 & \cellcolor{q1}0.01 & \cellcolor{q1}0.18 & \cellcolor{q1}0.12 & \cellcolor{q2}0.20 & \cellcolor{q1}0.10 & \cellcolor{q3}0.32 \\ 
		
	\midrule
	    \Flights & \cellcolor{q2}0.29 & \cellcolor{q3}0.30 & \cellcolor{q2}0.24 & \cellcolor{q3}0.34 & \cellcolor{q3}0.37 & \cellcolor{q1}0.18 & \cellcolor{q3}0.41 \\
	    
	    \CanParl & \cellcolor{q3}0.31 & \cellcolor{q1}0.01 & \cellcolor{q1}0.02 & \cellcolor{q1}0.08 & \cellcolor{q1}0.05 & \cellcolor{q1}0.01 & \cellcolor{q1}0.12 \\ 
	    
	    \USLegis & \cellcolor{q3}0.31 & \cellcolor{q1}0.02 & \cellcolor{q1}0.07 & \cellcolor{q2}0.21 & \cellcolor{q1}0.16 & \cellcolor{q1}-0.05 & \cellcolor{q1}0.10 \\ 
	    
	    \UNTrade & \cellcolor{q1}0.08 & \cellcolor{q1}0.03 & \cellcolor{q1}0.07 & \cellcolor{q1}0.07 & \cellcolor{q2}0.25 & \cellcolor{q1}-0.13 & \cellcolor{q1}0.05 \\ 
	    \UNVote & \cellcolor{q1}-0.02 & \cellcolor{q1}0.01 & \cellcolor{q1}0.01 & \cellcolor{q1}0.04 & \cellcolor{q1}0.07 & \cellcolor{q1}-0.13 & \cellcolor{q1}0.03 \\ 
	    \contact & \cellcolor{q3}0.59 & \cellcolor{q1}0.05 & \cellcolor{q1}0.01 & \cellcolor{q3}0.42 & \cellcolor{q1}0.17 & \cellcolor{q1}0.12 & \cellcolor{q2}0.28 \\
	\bottomrule
	\bottomrule
	\end{tabular}
\end{table*}

\begin{table*}[ht]
\caption{AU-ROC of dynamic link prediction in \textit{historical} negative sampling setting. The results report the mean over five runs with the standard deviations in parenthesis.}
\label{tab:au_roc_hist_ns}
\resizebox{\linewidth}{!}{%
\begin{tabular}{l c c c c c | c c }
	\toprule \toprule
		Data & JODIE & DyRep & TGAT & TGN & CAWN & $\text{\method}_\text{tw}$ & $\text{\method}_\infty$ \\ 
	\midrule
		\Wikipedia & 0.79 (0.010) & 0.79 (0.004) & 0.74 (0.006) & 0.84 (0.001) & 0.84 (0.069) & 0.77 (0.001) & 0.49 (0.001) \\ 
		\Reddit & 0.79 (0.005) & 0.80 (0.002) & 0.78 (0.002) & 0.81 (0.002) & 0.85 (0.035) & 0.77 (0.001) & 0.51 (0.000) \\ 
		\MOOC & 0.77 (0.041) & 0.80 (0.014) & 0.61 (0.046) & 0.85 (0.017) & 0.60 (0.368) & 0.60 (0.001) & 0.29 (0.001) \\
		\LastFM & 0.69 (0.057) & 0.70 (0.038) & 0.50 (0.003) & 0.77 (0.055) & 0.40 (0.120) & 0.76 (0.000) & 0.50 (0.000) \\
		\Enron & 0.62 (0.013) & 0.74 (0.017) & 0.53 (0.022) & 0.75 (0.040) & 0.51 (0.125) & 0.75 (0.002) & 0.48 (0.001) \\ 
		\SocialEvo & 0.83 (0.038) & 0.93 (0.002) & 0.78 (0.008) & 0.95 (0.004) & 0.56 (0.071) & 0.80 (0.000) & 0.55 (0.000) \\ 
		\UCI & 0.71 (0.091) & 0.44 (0.083) & 0.57 (0.009) & 0.72 (0.026) & 0.73 (0.095) & 0.69 (0.001) & 0.35 (0.004) \\ 
	\midrule
	    \Flights & 0.67 (0.013) & 0.66 (0.004) & 0.65 (0.004) & 0.66 (0.011) & 0.61 (0.070) & 0.71 (0.000) & 0.47 (0.000) \\
	    \CanParl & 0.45 (0.037) & 0.64 (0.032) & 0.71 (0.048) & 0.63 (0.056) & 0.86 (0.024) & 0.63 (0.001) & 0.27 (0.001) \\ 
	    \USLegis & 0.49 (0.057) & 0.69 (0.093) & 0.73 (0.044) & 0.68 (0.036) & 0.74 (0.187) & 0.68 (0.003) & 0.39 (0.003) \\ 
	    \UNTrade & 0.60 (0.010) & 0.59 (0.011) & 0.52 (0.010) & 0.61 (0.012) & 0.60 (0.020) & 0.81 (0.001) & 0.54 (0.000) \\
	    \UNVote & 0.70 (0.022) & 0.68 (0.007) & 0.51 (0.005) & 0.73 (0.025) & 0.65 (0.038) & 0.79 (0.001) & 0.53 (0.000) \\
	    \contact & 0.35 (0.002) & 0.51 (0.030) & 0.54 
	    (0.009) & 0.69 (0.033) & 0.71 (0.019) & 0.84 (0.000) & 0.54 (0.000) \\
	\bottomrule
	\bottomrule
	\end{tabular}
	}
\end{table*}

\begin{table*}[ht]
\centering
\caption{The AU-ROC loss of \textit{historical} compared to random negative sampling. The intensity of the color relates to the amount of loss.}
\label{tab:au_roc_hist_ns_difference}
\begin{tabular}{l c c c c c | c c }
	\toprule \toprule
		Data & JODIE & DyRep & TGAT & TGN & CAWN & $\text{\method}_\text{tw}$ & $\text{\method}_\infty$ \\ 
	\midrule
		\Wikipedia & 
		\cellcolor{q1}0.17 & \cellcolor{q1}0.16 & \cellcolor{q2}0.21 & \cellcolor{q1}0.14 & \cellcolor{q1}0.14 & \cellcolor{q1}0.10 & \cellcolor{q3}0.42 \\ 
		
		\Reddit & \cellcolor{q1}0.19 & \cellcolor{q1}0.18 & \cellcolor{q2}0.20 & \cellcolor{q1}0.18 & \cellcolor{q1}0.15 & \cellcolor{q1}0.14 & \cellcolor{q3}0.44 \\ 
		
		\MOOC & \cellcolor{q1}0.06 & \cellcolor{q1}0.02 & \cellcolor{q1}0.04 & \cellcolor{q1}0.07 & \cellcolor{q1}0.12 & \cellcolor{q1}0.01 & \cellcolor{q2}0.26 \\ 
		
		\LastFM & \cellcolor{q1}0.00 & \cellcolor{q1}0.01 & \cellcolor{q1}0.00 & \cellcolor{q1}-0.04 & \cellcolor{q3}0.57 & \cellcolor{q1}0.08 & \cellcolor{q3}0.33 \\ 
		
		\Enron & \cellcolor{q2}0.21 & \cellcolor{q1}0.08 & \cellcolor{q1}0.09 & \cellcolor{q1}0.12 & \cellcolor{q3}0.43 & \cellcolor{q1}0.12 & \cellcolor{q3}0.37 \\ 
		
		\SocialEvo & \cellcolor{q1}0.03 & \cellcolor{q1}-0.04 & \cellcolor{q1}0.00 & \cellcolor{q1}0.01 & \cellcolor{q1}0.11 & \cellcolor{q1}-0.12 & \cellcolor{q1}-0.01 \\ 
		
		\UCI & \cellcolor{q1}0.11 & \cellcolor{q1}0.00 & \cellcolor{q2}0.23 & \cellcolor{q1}0.16 & \cellcolor{q2}0.26 & \cellcolor{q1}0.07 & \cellcolor{q3}0.42 \\ 
		
	\midrule
	    \Flights & \cellcolor{q2}0.27 & \cellcolor{q2}0.28 & \cellcolor{q2}0.25 & \cellcolor{q3}0.32 & \cellcolor{q3}0.39 & \cellcolor{q1}0.13 & \cellcolor{q3}0.43 \\ 
	    
	    \CanParl & \cellcolor{q3}0.36 & \cellcolor{q1}0.00 & \cellcolor{q1}0.02 & \cellcolor{q1}0.07 & \cellcolor{q1}0.06 & \cellcolor{q1}0.01 & \cellcolor{q3}0.33 \\ 
	    
	    \USLegis & \cellcolor{q3}0.35 & \cellcolor{q1}0.01 & \cellcolor{q1}0.04 & \cellcolor{q1}0.15 & \cellcolor{q2}0.23 & \cellcolor{q1}-0.05 & \cellcolor{q2}0.20 \\ 
	    
	    \UNTrade & \cellcolor{q1}0.07 & \cellcolor{q1}0.03 & \cellcolor{q1}0.08 & \cellcolor{q1}0.07 & \cellcolor{q3}0.36 & \cellcolor{q1}-0.14 & \cellcolor{q1}0.09 \\ 
	    
	    \UNVote & \cellcolor{q1}-0.04 & \cellcolor{q1}0.00 & \cellcolor{q1}0.01 & \cellcolor{q1}0.02 & \cellcolor{q1}0.11 & \cellcolor{q1}-0.17 & \cellcolor{q1}0.05 \\ 
	    
	    \contact & \cellcolor{q3}0.64 & \cellcolor{q1}0.06 & \cellcolor{q1}0.02 & \cellcolor{q2}0.30 & \cellcolor{q2}0.25 & \cellcolor{q1}0.09 & \cellcolor{q2}0.33 \\
	    
	\bottomrule
	\bottomrule
	\end{tabular}
\end{table*}

\begin{table*}[ht]
\caption{AP of dynamic link prediction in \textit{inductive} negative sampling setting. The results report the mean over five runs with the standard deviations in parenthesis.}
\label{tab:ap_induc_ns}
\resizebox{\linewidth}{!}{%
\begin{tabular}{l c c c c c | c c }
	\toprule \toprule
		Data & JODIE & DyRep & TGAT & TGN & CAWN & $\text{\method}_\text{tw}$ & $\text{\method}_\infty$ \\ 
	\midrule
		\Wikipedia & 0.66 (0.027) & 0.69 (0.021) & 0.82 (0.007) & 0.87 (0.007) & 0.86 (0.045) & 0.46 (0.000) & 0.48 (0.000) \\ 
		\Reddit & 0.84 (0.005) & 0.85 (0.003) & 0.88 (0.002) & 0.88 (0.004) & 0.97 (0.018) & 0.47 (0.000) & 0.49 (0.000) \\ 
		\MOOC & 0.66 (0.020) & 0.63 (0.023) & 0.54 (0.016) & 0.77 (0.026) & 0.66 (0.277) & 0.42 (0.000) & 0.42 (0.000) \\
		\LastFM & 0.60 (0.030) & 0.63 (0.013) & 0.50 (0.000) & 0.67 (0.073) & 0.70 (0.34) & 0.46 (0.000) & 0.48 (0.000) \\
		\Enron & 0.59 (0.015) & 0.67 (0.016) & 0.57 (0.044) & 0.70 (0.019) & 0.57 (0.070) & 0.54 (0.000) & 0.54 (0.000) \\ 
		\SocialEvo & 0.72 (0.052) & 0.92 (0.004) & 0.79 (0.009) & 0.95 (0.005) & 0.60 (0.070) & 0.69 (0.000) & 0.55 (0.000) \\ 
		\UCI & 0.49 (0.013) & 0.54 (0.026) & 0.62 (0.008) & 0.69 (0.011) & 0.83 (0.063) & 0.43 (0.000) & 0.44 (0.000) \\ 
	\midrule
	    \Flights & 0.68 (0.022) & 0.65 (0.017) & 0.70 (0.011) & 0.69 (0.015) & 0.69 (0.069) & 0.47 (0.000) & 0.49 (0.000) \\
	    \CanParl & 0.48 (0.013) & 0.57 (0.009) & 0.67 (0.033) & 0.52 (0.037) & 0.85 (0.022) & 0.59 (0.001) & 0.55 (0.000) \\ 
	    \USLegis & 0.45 (0.020) & 0.64 (0.084) & 0.58 (0.052) & 0.53 (0.018) & 0.81 (0.130) & 0.65 (0.001) & 0.56 (0.000) \\ 
	    \UNTrade & 0.59 (0.020) & 0.62 (0.006) & 0.54 (0.030) & 0.63 (0.018) & 0.67 (0.036) & 0.56 (0.000) & 0.55 (0.001) \\ 
	    \UNVote & 0.67 (0.022) & 0.64 (0.011) & 0.51 (0.014) & 0.70 (0.022) & 0.76 (0.014) & 0.55 (0.000) & 0.53 (0.001) \\ 
	    \contact & 0.42 (0.001) & 0.51 (0.031) & 0.55 (0.006) & 0.58 (0.025) & 0.77 (0.015) & 0.49 (0.000) & 0.50 (0.000) \\
	\bottomrule
	\bottomrule
	\end{tabular}
	}
\end{table*}

\begin{table*}[ht]
\centering
\caption{The AP loss of \textit{inductive} compared to random negative sampling. The intensity of the color relates to the amount of loss.}
\label{tab:ap_induc_ns_difference}
\begin{tabular}{l c c c c c | c c }
	\toprule \toprule
		Data & JODIE & DyRep & TGAT & TGN & CAWN & $\text{\method}_\text{tw}$ & $\text{\method}_\infty$ \\ 
	\midrule
		\Wikipedia & \cellcolor{q2}0.28 & \cellcolor{q2}0.26 & \cellcolor{q1}0.14 & \cellcolor{q1}0.12 & \cellcolor{q1}0.13 & \cellcolor{q3}0.41 & \cellcolor{q3}0.41 \\ 
		
		\Reddit & \cellcolor{q1}0.11 & \cellcolor{q1}0.13 & \cellcolor{q1}0.10 & \cellcolor{q1}0.11 & \cellcolor{q1}0.03 & \cellcolor{q3}0.44 & \cellcolor{q3}0.46 \\ 
		
		\MOOC & \cellcolor{q1}0.12 & \cellcolor{q1}0.17 & \cellcolor{q1}0.06 & \cellcolor{q1}0.13 & \cellcolor{q1}0.09 & \cellcolor{q1}0.16 & \cellcolor{q1}0.11 \\ 
		
		\LastFM & \cellcolor{q1}0.08 & \cellcolor{q1}0.08 & \cellcolor{q1}0.00 & \cellcolor{q1}0.04 & \cellcolor{q2}0.28 & \cellcolor{q3}0.33 & \cellcolor{q2}0.30 \\ 
		
		\Enron & \cellcolor{q1}0.18 & \cellcolor{q1}0.12 & \cellcolor{q1}0.02 & \cellcolor{q1}0.15 & \cellcolor{q3}0.38 & \cellcolor{q2}0.030 & \cellcolor{q2}0.26 \\ 
		
		\SocialEvo & \cellcolor{q1}0.08 & \cellcolor{q1}-0.05 & \cellcolor{q1}-0.03 & \cellcolor{q1}-0.01 & \cellcolor{q1}0.12 & \cellcolor{q1}-0.08 & \cellcolor{q1}-0.03 \\ 
		
		\UCI & \cellcolor{q2}0.26 & \cellcolor{q1}-0.08 & \cellcolor{q1}0.17 & \cellcolor{q1}0.019 & \cellcolor{q1}0.16 & \cellcolor{q3}0.32 & \cellcolor{q3}0.33 \\ 
		
	\midrule
	    \Flights & \cellcolor{q2}0.25 & \cellcolor{q2}0.28 & \cellcolor{q1}0.19 & \cellcolor{q2}0.28 & \cellcolor{q3}0.30 & \cellcolor{q3}0.37 & \cellcolor{q3}0.40 \\ 
	    
	    \CanParl & \cellcolor{q2}0.27 & \cellcolor{q1}0.01 & \cellcolor{q1}0.01 & \cellcolor{q1}0.12 & \cellcolor{q1}0.10 & \cellcolor{q1}0.05 & \cellcolor{q1}0.06 \\ 
	    
	    \USLegis & \cellcolor{q3}0.31 & \cellcolor{q1}0.00 & \cellcolor{q1}0.12 & \cellcolor{q2}0.24 & \cellcolor{q1}0.17 & \cellcolor{q1}-0.06 & \cellcolor{q1}-0.01 \\  
	    
	    \UNTrade & \cellcolor{q1}0.05 & \cellcolor{q1}-0.01 & \cellcolor{q1}0.04 & \cellcolor{q1}0.01 & \cellcolor{q3}0.31 & \cellcolor{q1}0.04 & \cellcolor{q1}0.02 \\  
	    
	    \UNVote & \cellcolor{q1}-0.02 & \cellcolor{q1}0.01 & \cellcolor{q1}0.00 & \cellcolor{q1}0.01 & \cellcolor{q1}0.06 & \cellcolor{q1}0.03 & \cellcolor{q1}0.02 \\ 
	    
	    \contact & \cellcolor{q3}0.57 & \cellcolor{q1}0.05 & \cellcolor{q1}0.03 & \cellcolor{q3}0.41 & \cellcolor{q1}0.20 & \cellcolor{q3}0.40 & \cellcolor{q3}0.30\\
	\bottomrule
	\bottomrule
	\end{tabular}
\end{table*}

\begin{table*}[ht]
\caption{AU-ROC of dynamic link prediction in \textit{inductive} negative sampling setting. The results report the mean over five runs with the standard deviations in parenthesis.}
\label{tab:au_roc_induc_ns}
\resizebox{\linewidth}{!}{%
\begin{tabular}{l c c c c c | c c }
	\toprule \toprule
		Data & JODIE & DyRep & TGAT & TGN & CAWN & $\text{\method}_\text{tw}$ & $\text{\method}_\infty$ \\ 
	\midrule
		\Wikipedia & 0.66 (0.017) & 0.67 (0.016) & 0.79 (0.006) & 0.82 (0.004) & 0.80 (0.067) & 0.40 (0.001) & 0.43 (0.000) \\ 
		\Reddit & 0.81 (0.003) & 0.82 (0.002) & 0.86 (0.001) & 0.85 (0.003) & 0.96 (0.024) & 0.43 (0.000) & 0.47 (0.000) \\ 
		\MOOC & 0.67 (0.024) & 0.63 (0.028) & 0.55 (0.012) & 0.77 (0.027) & 0.60 (0.366) & 0.19 (0.000) & 0.22 (0.000) \\
		\LastFM & 0.59 (0.020) & 0.61 (0.014) & 0.50 (0.003) & 0.65 (0.071) & 0.59 (0.53) & 0.41 (0.000) & 0.45 (0.000) \\
		\Enron & 0.63 (0.015) & 0.68 (0.015) & 0.58 (0.036) & 0.71 (0.021) & 0.42 (0.096) & 0.52 (0.000) & 0.53 (0.000) \\ 
		\SocialEvo & 0.82 (0.036) & 0.93 (0.002) & 0.80 (0.009) & 0.95 (0.004) & 0.50 (0.046) & 0.77 (0.000) & 0.59 (0.000) \\ 
		\UCI & 0.55 (0.015) & 0.54 (0.029) & 0.59 (0.010) & 0.62 (0.014) & 0.78 (0.087) & 0.29 (0.000) & 0.31 (0.000) \\ 
	\midrule
	    \Flights & 0.69 (0.020) & 0.67 (0.011) & 0.70 (0.007) & 0.70 (0.014) & 0.68 (0.071) & 0.38 (0.000) & 0.44 (0.000) \\
	    \CanParl & 0.50 (0.028) & 0.63 (0.021) & 0.71 (0.036) & 0.58 (0.047) & 0.79 (0.030) & 0.54 (0.001) & 0.49 (0.002) \\ 
	    \USLegis & 0.50 (0.046) & 0.71 (0.097) & 0.68 (0.038) & 0.64 (0.028) & 0.72 (0.188) & 0.69 (0.001) & 0.60 (0.002) \\ 
	    \UNTrade & 0.64 (0.017) & 0.63 (0.019) & 0.58 (0.042) & 0.68 (0.018) & 0.52 (0.053) & 0.57 (0.000) & 0.57 (0.000) \\ 
	    \UNVote & 0.71 (0.024) & 0.69 (0.009) & 0.53 (0.022) & 0.77 (0.026) & 0.67 (0.024) & 0.58 (0.000) & 0.56 (0.000) \\ 
	    \contact & 0.41 (0.003) & 0.52 (0.027) & 0.52 (0.007) & 0.70 (0.023) & 0.67 (0.021) & 0.48 (0.000) & 0.49 (0.000) \\
	\bottomrule
	\bottomrule
	\end{tabular}
	}
\end{table*}

\begin{table*}[ht]
\centering
\caption{The AU-ROC loss of \textit{inductive} compared to random negative sampling. The intensity of the color relates to the amount of loss.}
\label{tab:au_roc_induc_ns_difference}
\begin{tabular}{l c c c c c | c c }
	\toprule \toprule
		Data & JODIE & DyRep & TGAT & TGN & CAWN & $\text{\method}_\text{tw}$ & $\text{\method}_\infty$ \\ 
	\midrule
		\Wikipedia & \cellcolor{q3}0.30 & \cellcolor{q2}0.28 & \cellcolor{q1}0.17 & \cellcolor{q1}0.17 & \cellcolor{q1}0.19 & \cellcolor{q3}0.47 & \cellcolor{q3}0.47 \\ 
		
		\Reddit & \cellcolor{q1}0.16 & \cellcolor{q1}0.16 & \cellcolor{q1}0.12 & \cellcolor{q1}0.14 & \cellcolor{q1}0.04 & \cellcolor{q3}0.49 & \cellcolor{q3}0.49 \\ 
		
		\MOOC & \cellcolor{q1}0.16 & \cellcolor{q1}0.19 & \cellcolor{q1}0.10 & \cellcolor{q1}0.15 & \cellcolor{q1}0.12 & \cellcolor{q3}0.42 & \cellcolor{q3}0.33 \\ 
		
		\LastFM & \cellcolor{q1}0.10 & \cellcolor{q1}0.10 & \cellcolor{q1}0.00 & \cellcolor{q1}0.08 & \cellcolor{q3}0.39 & \cellcolor{q3}0.43 & \cellcolor{q3}0.39 \\ 
		
		\Enron & \cellcolor{q1}0.19 & \cellcolor{q1}0.14 & \cellcolor{q1}0.04 & \cellcolor{q1}0.17 & \cellcolor{q3}0.52 & \cellcolor{q3}0.35 & \cellcolor{q3}0.32 \\ 
		
		\SocialEvo & \cellcolor{q1}0.04 & \cellcolor{q1}-0.03 & \cellcolor{q1}-0.02 & \cellcolor{q1}0.01 & \cellcolor{q1}0.17 & \cellcolor{q1}-0.10 & \cellcolor{q1}-0.06 \\ 
		
		\UCI & \cellcolor{q2}0.28 & \cellcolor{q1}-0.10 & \cellcolor{q2}0.22 & \cellcolor{q2}0.26 & \cellcolor{q2}0.21 & \cellcolor{q3}0.47 & \cellcolor{q3}0.47 \\ 
		
	\midrule
	    \Flights & \cellcolor{q2}0.26 & \cellcolor{q2}0.27 & \cellcolor{q2}0.20 & \cellcolor{q2}0.28 & \cellcolor{q3}0.32 & \cellcolor{q3}0.46 & \cellcolor{q3}0.46 \\ 
	    
	    \CanParl & \cellcolor{q3}0.31 & \cellcolor{q1}0.01 & \cellcolor{q1}0.02 & \cellcolor{q1}0.13 & \cellcolor{q1}0.13 & \cellcolor{q1}0.11 & \cellcolor{q1}0.11 \\ 
	    
	    \USLegis & \cellcolor{q3}0.35 & \cellcolor{q1}-0.01 & \cellcolor{q1}0.08 & \cellcolor{q1}0.19 & \cellcolor{q2}0.24 & \cellcolor{q1}-0.06 & \cellcolor{q1}-0.01 \\ 
	    
	    \UNTrade & \cellcolor{q1}0.03 & \cellcolor{q1}-0.01 & \cellcolor{q1}0.02 & \cellcolor{q1}0.00 & \cellcolor{q3}0.44 & \cellcolor{q1}0.09 & \cellcolor{q1}0.05 \\ 
	    
	    \UNVote & \cellcolor{q1}-0.05 & \cellcolor{q1}-0.01 & \cellcolor{q1}-0.02 & \cellcolor{q1}-0.02 & \cellcolor{q1}0.08 & \cellcolor{q1}0.04 & \cellcolor{q1}0.03 \\ 
	    
	    \contact & \cellcolor{q3}0.58 & \cellcolor{q1}0.05 & \cellcolor{q1}0.04 & \cellcolor{q2}0.29 & \cellcolor{q2}0.29 & \cellcolor{q3}0.48 & \cellcolor{q3}0.38\\
	\bottomrule
	\bottomrule
	\end{tabular}
\end{table*}


\subsection{More Discussion on Historical and Inductive Negative Sampling}
\label{subsec:historical_vs_inductive_NS}

\changed{
In Section \ref{sec:negative_sampling}, we explain different negative sampling strategies and mention that in case of historical and inductive negative sampling, if there are not enough historical or negative edges to sample from, we randomly sample negative edges to preserve dataset balance during the test phase.
Here, we elaborate more on the statistics of the negative samples selected during the test phase for each dataset.
}

\changed{
Specifically, in the historical negative sampling setting, all considered datasets have enough historical edges to sample from during the test phase. 
Thus, there is no need to sample random edges. The detailed statistics of the number of random versus historical negative edges during the test phase are given in \tableRef{tab:test_stat}, second and third column.
}


\changed{
In the inductive negative sampling setting, at the first batches of the initial timestamps of the test phase, there are not enough inductive negative edges. 
Therefore, some random edges should be selected to compose a balanced dataset. 
As an example of such a dataset, we can consider \SocialEvo whose \TET plot is illustrated in \figureRef{fig:TET_SocialEvo}.
We can observed that this dataset does not have enough inductive edges, therefore the majority of the negative edges during the test phase are randomly selected.
This also affects the performance change for \SocialEvo dataset reported in \figureRef{fig:delta_induc}.
Because a lot of negative edges are sill selected randomly, we observe a less severe performance drop for \SocialEvo in \figureRef{fig:delta_induc}.
The detailed statistics of the number of random versus inductive negative edges in the test phase are given in \tableRef{tab:test_stat}, fourth and fifth column.
}


\begin{table*}[t]
\caption{\changed{Statistics of negative edges used during the test phase in historical or inductive NS setting.
}}\vspace{-5pt}
\label{tab:test_stat}
  \resizebox{\linewidth}{!}{%
  \begin{tabular}{l | c c | c c | c }
  \toprule
  \toprule
  \multirow{2}{*}{Dataset} & \multicolumn{2}{c|}{\textbf{Historical} NS} & \multicolumn{2}{c|}{\textbf{Inductive} NS} & \multirow{2}{*}{\# Total Negative Edges} \\
   & \# Random & \# Historical & \# Random & \# Inductive &  \\
  \midrule
  \Wikipedia    & 0     & 23,621  & 1,018     & 22,603   & 23,621    \\
  \Reddit       & 0     & 100,867 & 0         & 100,867  & 100,867   \\
  \MOOC         & 0     & 61,763  & 351       & 61,412   & 61,768    \\
  \LastFM       & 0     & 193,966 & 0         & 193,966  & 193,966   \\
  \Enron        & 0     & 18,785  & 3,689     & 15,096   & 18,785    \\
  \SocialEvo    & 0     & 314,924 & 268,958   & 45,966   & 314,924   \\
  \UCI          & 0     & 8,976   & 402       & 8,574    & 8,976     \\
  \midrule
  \Flights      & 0     & 287,824 & 18,800    & 269,024  & 287,824   \\
  \CanParl      & 0     & 10,113  & 7200      & 2,913    & 10,113    \\
  \USLegis      & 0     & 9,804   & 5,000     & 4,804    & 9,804     \\
  \UNTrade      & 0     & 61,595  & 20,800    & 40,795   & 61,595    \\
  \UNVote       & 0     & 155,119 & 73,167    & 81,952   & 155,119   \\
  \contact      & 0     & 363,780 & 5,227     & 358,553  & 363,780 \\
  \bottomrule
  \bottomrule
  \end{tabular}
  }
\vspace{-5pt}
\end{table*}

\begin{wrapfigure}{R}{0.65\textwidth}
\centering
\includegraphics[width=0.6\textwidth]{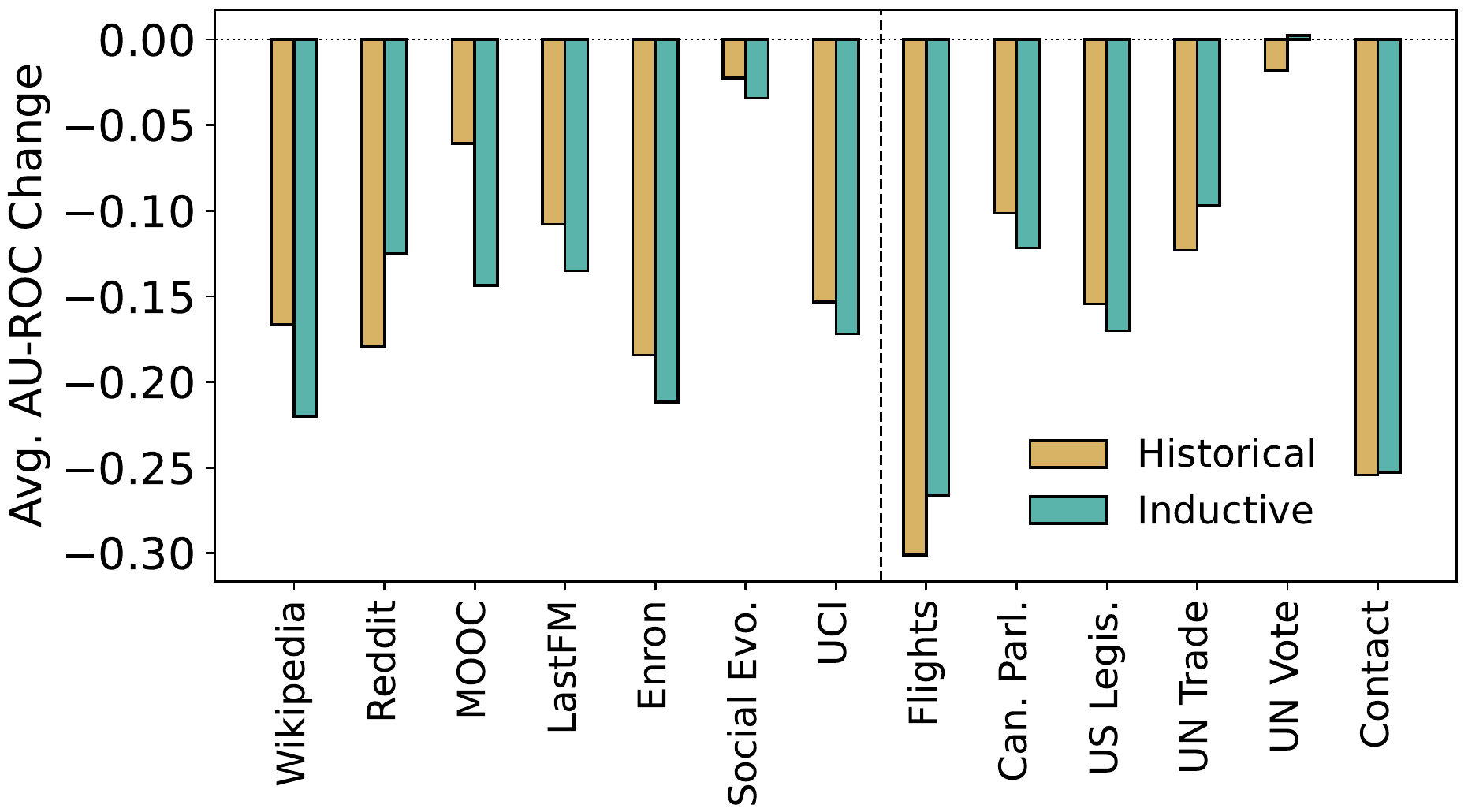}
\caption{Average AU-ROC change of SOTA methods for different NS strategies compared to random NS. The impact of moving to historical or inductive NS varies across datasets.}
\label{fig:data_perspective}
\vspace{-10pt}
\end{wrapfigure}

\figureRef{fig:data_perspective} demonstrates the average change in performance with historical and inductive NS across different SOTA methods. 
In general, the decrease is at least 10 percentage points. 
The new \Flights and \contact datasets are particularly challenging, with more than 25 percentage points average loss when comparing historical or inductive NS with random NS.
In case of the \Flights network, this can be interpreted as the models struggle to correctly predict whether a flight that happened in the past will happen again.

\newpage
\section{Broader Impact} 
\label{app:impact}
We expect this work to have a major impact on the fundamental as well as applied dynamic graph research.

Essentially, high-quality datasets from diverse domains play undeniable roles in advancement of research (e.g., OGB~\cite{hu2020open} or ImageNet~\cite{deng2009imagenet}).
By contributing 5 new datasets from less explored real-world domains, we aim to enrich available datasets for dynamic graph learning tasks, and facilitate the development of novel dynamic graph models.

In addition, our proposed dynamic graph visualization techniques (i.e., \TEA and \TET plot) together with the defined indices (i.e., novelty, reoccurrence, and surprise index) provide comprehensive summary of datasets characteristics.
\method also provides a simple yet strong baseline for the dynamic link prediction task that future learning models can easily compare against.
Additionally, our investigation on the impact of negative sampling in dynamic graphs leads to more robust evaluation setup for the dynamic link prediction task and facilitates methodological advancement in dynamic graph ML.

Since dynamic link prediction has many applications in different domains, such as recommendation systems, academic graphs, computational finance, etc., we expect this work to facilitate the development of applied methods in different domains as well.

\par{\textbf{Potential Negative Impact}}

In this work, we have investigated several dynamic graph datasets that are under study in dynamic graph research. One potential negative impact is that future research may narrow down their study to these datasets. We aim to regularly update the datasets with the input from the community to prevent this issue.

Additionally, improving link prediction can be associated with several potential negative use cases such as user profiling.
While our work does not directly lead to such negative impacts, being aware of such impacts is important and appropriate precautions should be considered.

\section{Limitations} 
\label{app:limit}
We consider two main limitations for this work:

First, in the current evaluation setup, there is a single point split for past and future links, which is the current common practice. It might be more relevant to consider alternative settings where temporal information plays a stronger role, from splitting in more time points to predict the exact time of an edge. 

Second, we have only considered the transductive setting where all nodes are seen during training, since this is the only setup that we could easily check for memorization. The baseline and historic negative sampling strategy proposed here are only considered in the transductive setting. 

In addition to these two main limitations, we only considered the dynamic link prediction task and leave the exploration of similar concepts in the related node classification task in dynamic graphs as future work.



\section{Maintenance Plan}
\label{subsec:maintenance}

\changed{
To provide an easy to use, robust and diverse benchmark for dynamic link prediction, we will be maintaining a code repository at \url{https://github.com/fpour/DGB.git} to run the experiments and benchmarks. The relevant datasets will be maintained at \url{https://zenodo.org/record/7008205\#.Yv_a_3bMJPZ}. These code and data repositories will be updated regularly to include more datasets and benchmarks as they become available. We also plan to increase accessibility of the benchmark by adding more documentations and tutorials and providing a simple command for python package manager.
}

\end{document}